\documentclass{article}

\usepackage{arxiv}

\usepackage[utf8]{inputenc} 
\usepackage[T1]{fontenc}    
\usepackage{hyperref}       
\usepackage{url}            
\usepackage{booktabs}       
\usepackage{amsfonts}       
\usepackage{nicefrac}       
\usepackage{microtype}      
\usepackage{lipsum}		
\usepackage{graphicx}
\usepackage{natbib}
\usepackage{doi}

\usepackage{amssymb}
\usepackage{subcaption}
\usepackage{xcolor}

\usepackage{algorithmic}
\usepackage{algorithm}

\newcommand{\edit}[1]{\textcolor{black}{#1}}
	
\usepackage{color, colortbl}
\definecolor{Gray}{gray}{0.9}
\usepackage{amsmath}

\usepackage{framed}
\usepackage{multicol}
\usepackage{nomencl}

\renewcommand*\nompreamble{\begin{multicols}{2}}
\renewcommand*\nompostamble{\end{multicols}}

\usepackage{etoolbox}
\renewcommand\nomgroup[1]{%
  \item[\bfseries
  \ifstrequal{#1}{A}{ }{%
  \ifstrequal{#1}{B}{Symbols in the equations}{%
  \ifstrequal{#1}{C}{Other Symbols}{}}}%
]}
\makenomenclature

\title{Health Index Estimation Through Integration of General Knowledge with Unsupervised Learning}

\author{ \href{https://orcid.org/0009-0002-7945-6053}{\includegraphics[scale=0.06]{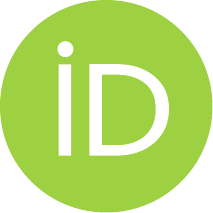}\hspace{1mm}Kristupas Bajarunas} \\
	Faculty of Aerospace Engineering,\\
	Delft University of Technology\\
	HS 2926 Delft, The Netherlands\\
	\texttt{k.v.b.bajarunas@tudelft.nl} \\
	\And
	{Marcia L. Baptista} \\
	Faculty of Aerospace Engineering,\\
	Delft University of Technology\\
	HS 2926 Delft, The Netherlands\\
	\texttt{m.lbaptista@tudelft.nl} \\
	\And
	{Kai Goebel} \\
	SRI International,\\
	CA 94304 Palo Alto, United States\\
	\texttt{kai.goebel@sri.com} \\
 	\And
	{Manuel Arias Chao} \\
	Faculty of Aerospace Engineering,\\
	Delft University of Technology\\
	HS 2926 Delft, The Netherlands\\
	\texttt{m.a.c.ariaschao@tudelft.nl} \\
 }

\renewcommand{\shorttitle}{\textit{arXiv} Template}

\hypersetup{
pdftitle={A template for the arxiv style},
pdfsubject={q-bio.NC, q-bio.QM},
pdfauthor={Kristupas ~Bajarunas},
pdfkeywords={Health Index, Prognostics,Unsupervised Learning},
}

\begin{document}
\maketitle

\begin{abstract}
Accurately estimating a Health Index (HI) from condition monitoring data (CM) is essential for reliable and interpretable prognostics and health management (PHM) in complex systems. In most scenarios, complex systems operate under varying operating conditions and can exhibit different fault modes, making unsupervised inference of an HI from CM data a significant challenge. Hybrid models combining prior knowledge about degradation with deep learning models have been proposed to overcome this challenge. However, previously suggested hybrid models for HI estimation usually rely heavily on system-specific information, limiting their transferability to other systems. In this work, we propose an unsupervised hybrid method for HI estimation that integrates general knowledge about degradation into the convolutional autoencoder's model architecture and learning algorithm, enhancing its applicability across various systems. The effectiveness of the proposed method is demonstrated in two case studies from different domains: turbofan engines and lithium batteries. The results show that the proposed method outperforms other competitive alternatives, including residual-based methods, in terms of HI quality and their utility for Remaining Useful Life (RUL) predictions. The case studies also highlight the comparable performance of our proposed method with a supervised model trained with HI labels.
\end{abstract}


\section{Introduction}
\label{Introduction}

\nomenclature[A,01]{HI}{health index}
\nomenclature[A,02]{PHM}{prognostics and health management}
\nomenclature[A,03]{CM}{condition monitoring}
\nomenclature[A,04]{RUL}{remaining useful life}
\nomenclature[A,05]{SL}{supervised learning}
\nomenclature[A,06]{UL}{unsupervised learning}
\nomenclature[A,07]{RM}{residual method}
\nomenclature[A,08]{AE}{AutoEncoder}
\nomenclature[A,09]{PCA}{principal component analysis}
\nomenclature[A,10]{CNN}{convolutional neural network}
\nomenclature[A,11]{MAE}{mean absolute error}
\nomenclature[A,12]{RMSE}{root mean squared error}
\nomenclature[A,13]{MAPE}{mean absolute percentage error}
\nomenclature[A,14]{Mon}{monotonicity}
\nomenclature[A,15]{Tren}{trendability}
\nomenclature[A,16]{Prog}{prognosability}
\nomenclature[A,17]{MutInf}{mutual information score}
\nomenclature[A,18]{CMAPSS}{commercial modular aero-propulsion system simulation}
\nomenclature[A,19]{SCM}{structural causal model}
\nomenclature[A,20]{DAG}{directed acyclic graph}
\nomenclature[A,19]{ANM}{additive noise model}

\nomenclature[B,01]{$X$}{sensor readings}
\nomenclature[B,02]{$W$}{operating conditions}
\nomenclature[B,03]{$Z$}{degradation (representation)}
\nomenclature[B,04]{$T$}{cycle number}
\nomenclature[B,05]{$u$}{unit of a fleet}
\nomenclature[B,06]{$m$}{observations}
\nomenclature[B,07]{$p$}{number of sensors}
\nomenclature[B,08]{$k$}{number of operating conditions}
\nomenclature[B,09]{$C$}{correlation constraint}
\nomenclature[B,10]{$NG$}{negative gradient constraint}
\nomenclature[B,11]{$F$}{functional constraint}

\begin{table*}[!t]   
\begin{framed}

\printnomenclature

\end{framed}
\end{table*}

Understanding the health condition of complex systems is an important step in prognostics and health management (PHM) \cite{LEI2018799,liu2013data}. A Health Index (HI) represents the system's health state over time or usage on a scale from 1 (perfect health) to 0 (failure) and, therefore, provides a clear and interpretable measure of degradation. HIs are also instrumental in predicting remaining useful life (RUL). For instance, HIs can be integrated into prognostic models by matching HI patterns with known failure times \cite{wang2008similarity,wang2012generic,yu2020improved,liu2019remaining}, or extrapolated until the failure threshold \cite{guo2017recurrent,kumar2022state,zhang2024health} for RUL predictions.

Different data-driven approaches have been proposed for estimating HI from condition monitoring (CM) data, but many of these approaches rely extensively on labeled data. For instance, when dealing with datasets containing HI labels, the utilization of supervised models is prevalent \cite{roman2021machine}. Another common strategy is the residual technique, where models are trained to recognize a system's normal behavior using health state labels, subsequently identifying the HI by analyzing reconstruction errors \cite{ye2021health} \edit{\cite{de2023developing,lovberg2021remaining}}. However, for complex systems, the challenge lies in obtaining representative labeled data, which can be costly or unfeasible in industrial contexts. This limitation has motivated a growing interest in unsupervised learning methods for HI estimation, circumventing the need for labeled datasets.


\edit{To address the difficulty of dealing with unlabeled data, researchers have proposed hybrid unsupervised models combining data-driven models with prior knowledge about the system for HI estimation. For instance, Biggio et. al. \cite{biggio2023ageing} leverage a battery simulator for training a transformer architecture with synthethic data of degraded system dynamics, allowing their model to uncover degradation patterns from real-world experiments. Alternatively, Guo et. al. \cite{guo2022health} combine a data-driven HI with knowledge-based HI to model power transformer degradation. Nonetheless, a characteristic of many current hybrid models is their dependence on system-specific knowledge (e.g., detailed simulators of degraded system dynamics), limiting their applicability to other systems exhibiting diverse degradation patterns. Moreover, as pointed out in recent review work \cite{li2024review}, most models rely on a single strategy to integrate data and prior knowledge, potentially limiting their effectiveness. These gaps hinder the broader application of hybrid models and emphasize the need for more general hybrid models that accommodate a wide array of complex systems \cite{Hagmeyer2022}}.

In this paper, we build upon our initial study \cite{bajarunas2023unsupervised}, where we showcased the feasibility of inferring HIs for turbofan engines with a hybrid unsupervised method. In this current research, we expand the methodology with the primary objective of demonstrating the generalization of our method across various complex systems. To this end, we address the following research question: 
\textit{How can knowledge about degradation be incorporated into an unsupervised hybrid method for HI estimation applicable to diverse complex systems?}


To achieve the intended generalization, we propose to integrate general domain knowledge about the HI problem into the method using multiple hybridization strategies. Specifically, we postulate that there are common fundamental degradation characteristics at a certain level of abstraction that are informative and, hence, transferable across a range of complex systems. For instance, degradation characteristics of multiple complex systems exhibit a fast wear period, followed by a period of almost steady decline, which is ultimately followed by another period of faster wear towards the end of life.
Additionally, we hypothesize that  an understanding of known causal relationships can be leveraged for more accurate HI estimation

\begin{figure}[h!]
    \centering
    \includegraphics[width = \textwidth]{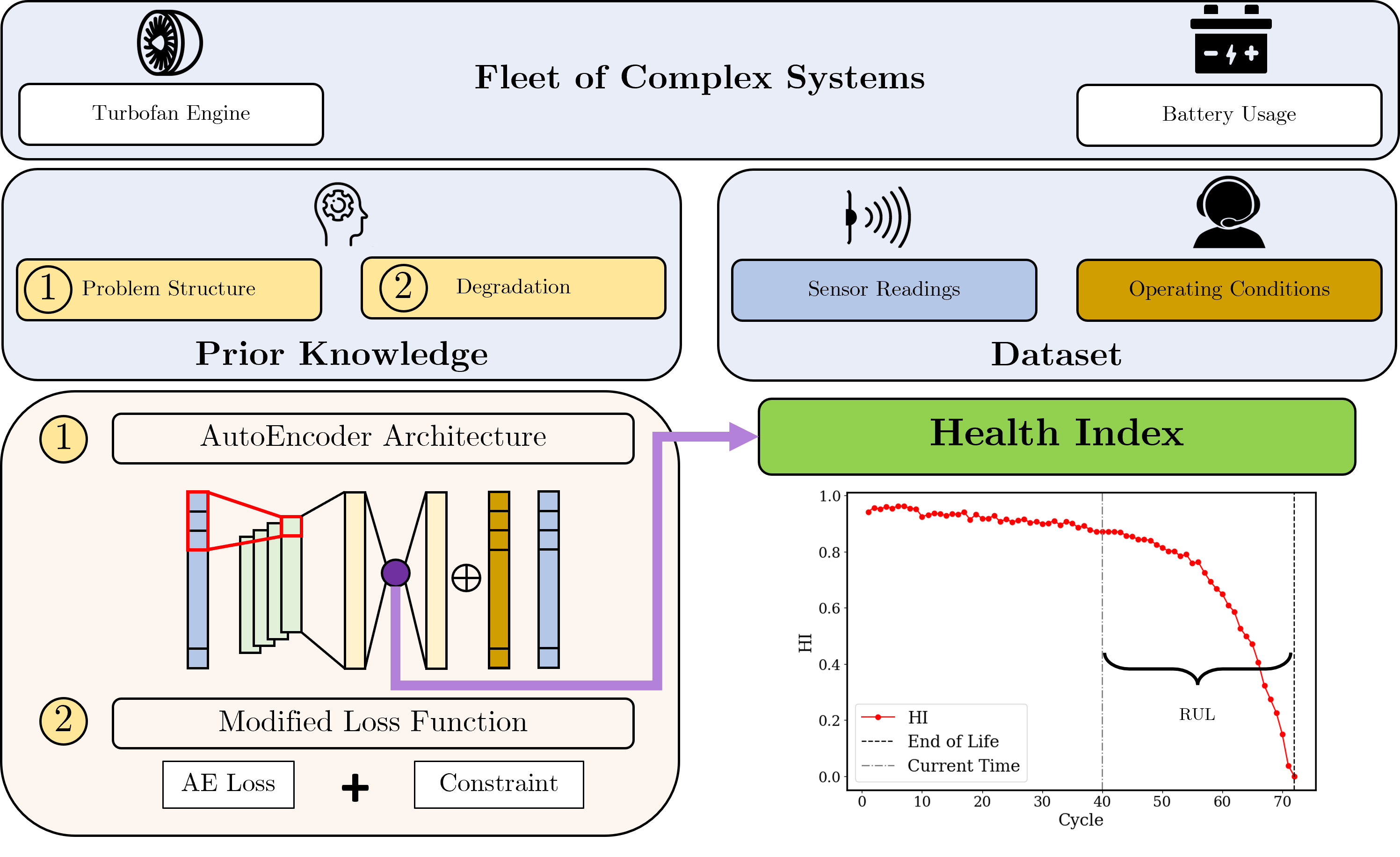}
    \caption{Overview of the unsupervised hybrid method for HI estimation that relies primarily on general knowledge about degradation.}
    \label{fig:graphicalabstract}
\end{figure}

In line with these hypotheses, we propose an unsupervised hybrid method for HI estimation with two distinct design features: 1) a novel network architecture of a convolutional AutoEncoder (AE) preserving the causal relationships among sensor readings, operating conditions, and degradation within complex systems, and 2) the incorporation of soft constraints within the loss function derived from general knowledge of the degradation process, guiding the AE to infer degradation in its latent space. Figure \ref{fig:graphicalabstract} provides an overview of our proposed HI estimation methodology.

To demonstrate the intended generalization, our experimental analysis investigates two distinct case studies, turbofan engines and Li-ion batteries. Our proposed method is thoroughly compared against two alternative methodologies: a residual-based method and a supervised model. This comparison encompasses scenarios both within and out of distribution. \edit{For reproducibility and future research purposes, the complete code reproducing this study is released in open-source at \url{https://github.com/KBaja/UnsupervisedHI}.}

The main contributions of this study are as follows:

\begin{enumerate}
    \item \edit{We propose a novel hybrid unsupervised method for HI estimation that relies on general knowledge about degradation and combines multiple hybridization techniques. We demonstrate that our model can accurately estimate the HI of various systems (turbofan engines and batteries) with distinct degradation patterns. }

    \item \edit{We provide an extensive comparative analysis involving supervised, residual, and unsupervised methods for HI estimation, enabling a quantified assessment of their respective performances. The outcomes highlight the superiority of our proposed unsupervised method over the residual method, positioning it at par with the supervised model. }

    \item \edit{We evaluate different forms of general knowledge options for integration into our model to be used as constraints in the latent space of an AE: monotonicity, negative gradient, and functional HI.}


        
\end{enumerate}




    

The remainder of this paper is organized as follows: Section \ref{sec:Knowledge} presents background information about general degradation dynamics. Section \ref{Background} presents the problem formulation and related work. Section \ref{Methodology} proposes the unsupervised hybrid model, while Section \ref{Case Studies} presents the case studies and the training set-up. In Section \ref{Results}, the results are presented, followed by a discussion in Section \ref{Discussion}, and the conclusion in  Section \ref{Conclusion}.

\section{General Domain Knowledge about Degradation}
\label{sec:Knowledge}
\edit{
While system-specific knowledge can be valuable for HI estimation, it often limits the generalizability of the approach. This section explores the concept of \textbf{general knowledge} about degradation, referring to domain knowledge that applies broadly to complex systems exhibiting degradation. In particular, we introduce two key examples of general knowledge utilized in this paper: the causal structure of condition monitoring data and degradation dynamics.}

\edit{
In the context of a complex system, the causal structure describes the underlying network of cause-and-effect relationships that link the various components together. Uncovering the causal structure is crucial for understanding how a complex system functions, responds to external influences and evolves over time. It involves identifying the key components, their interactions, and the mechanisms through which they influence each other, as well as the potential feedback loops and non-linear dynamics that can emerge from these interactions.}

\edit{
Knowledge of the underlying cause-and-effect relationships within a complex system can provide valuable insights into degradation processes. \footnote{Under the hypothesis that signatures of faults can be observed in the condition monitoring data.}. This knowledge, often regarded as general knowledge  \cite{si2015adaptive,nouri2022assessment,alves18group}, is based on a general understanding of which variables (causes) directly influence degradation and which variables (effects) are themselves affected by degradation. For instance, in a battery system, increasing the ambient temperature (cause) can accelerate the degradation of battery capacity (effect). From another perspective, a decrease in capacity (cause) can lead to faster constant current charging times (effect).}

Beyond understanding the cause-and-effect relationships within a system, general knowledge also extends to the concept of degradation dynamics. In complex systems, degradation often unfolds gradually, with failures evolving over time rather than occurring abruptly \cite{smith2021reliability}. Most failure modes stem from an underlying degradation process, where gradual deterioration eventually reveals weaknesses that can lead to system failure \cite{meeker2022statistical}. Degradation in complex systems manifests itself in various forms, ranging from observable changes in physical components like crack growth to subtler alterations affecting system dynamics and performance degradation, such as changes in battery output voltage. Despite the diverse manifestations of degradation, consistent patterns emerge across different systems and their degradation dynamics.

\begin{figure}[h!]
    \centering
    \includegraphics[scale = 0.5]{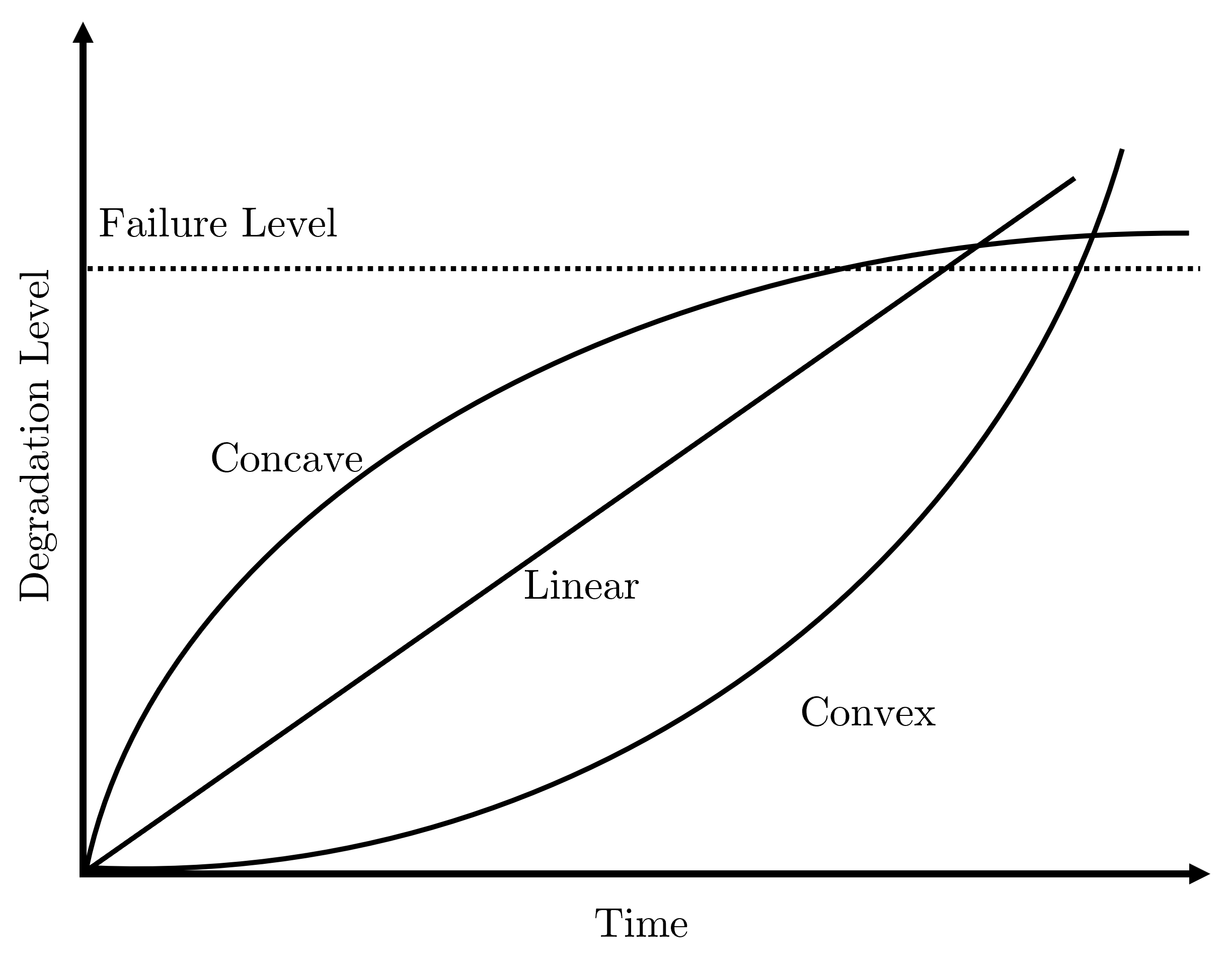}
    \caption{Possible shapes for univariate health degradation curves (adapted from \cite{meeker2022statistical})}
    \label{fig:Example Degradation}
\end{figure}

Figure \ref{fig:Example Degradation} illustrates three common temporal evolutions of degradation: linear, convex, and concave. \edit{While real-world systems may exhibit a combination of these patterns, this figure showcases each in isolation for clarity.}. The horizontal line on the degradation scale represents the failure threshold. Linear degradation involves a steady increase in degradation over time, 
such as the wear of automobile tire treads, \edit{which appears linear over a certain time.} Convex degradation entails an accelerating rate of degradation increase, as seen in crack growth scenarios. Conversely, concave degradation entails an increase in degradation over time at a diminishing rate, such as the growth of chlorine-copper compounds in printed circuit boards \cite{meeker2022statistical}.

At a certain level of abstraction, fundamental degradation characteristics remain consistent across various system dynamics \cite{Saxena2008}. For instance, degradation is typically monotonic or non-decreasing, as depicted in Figure \ref{fig:Example Degradation}, which may be expressed mathematically through positive first differences between observations or positive gradients with respect to time. Even though some of these commonalities may not universally apply to all real-world systems (e.g., the apparent short-term capacity recovery of batteries during no-load periods), we anticipate their persistence in the long-term degradation process.

\section{Problem Formulation and Related Work}
\label{Background}
\begin{figure}[h!]%
    \centering
    \includegraphics[scale = 0.5]{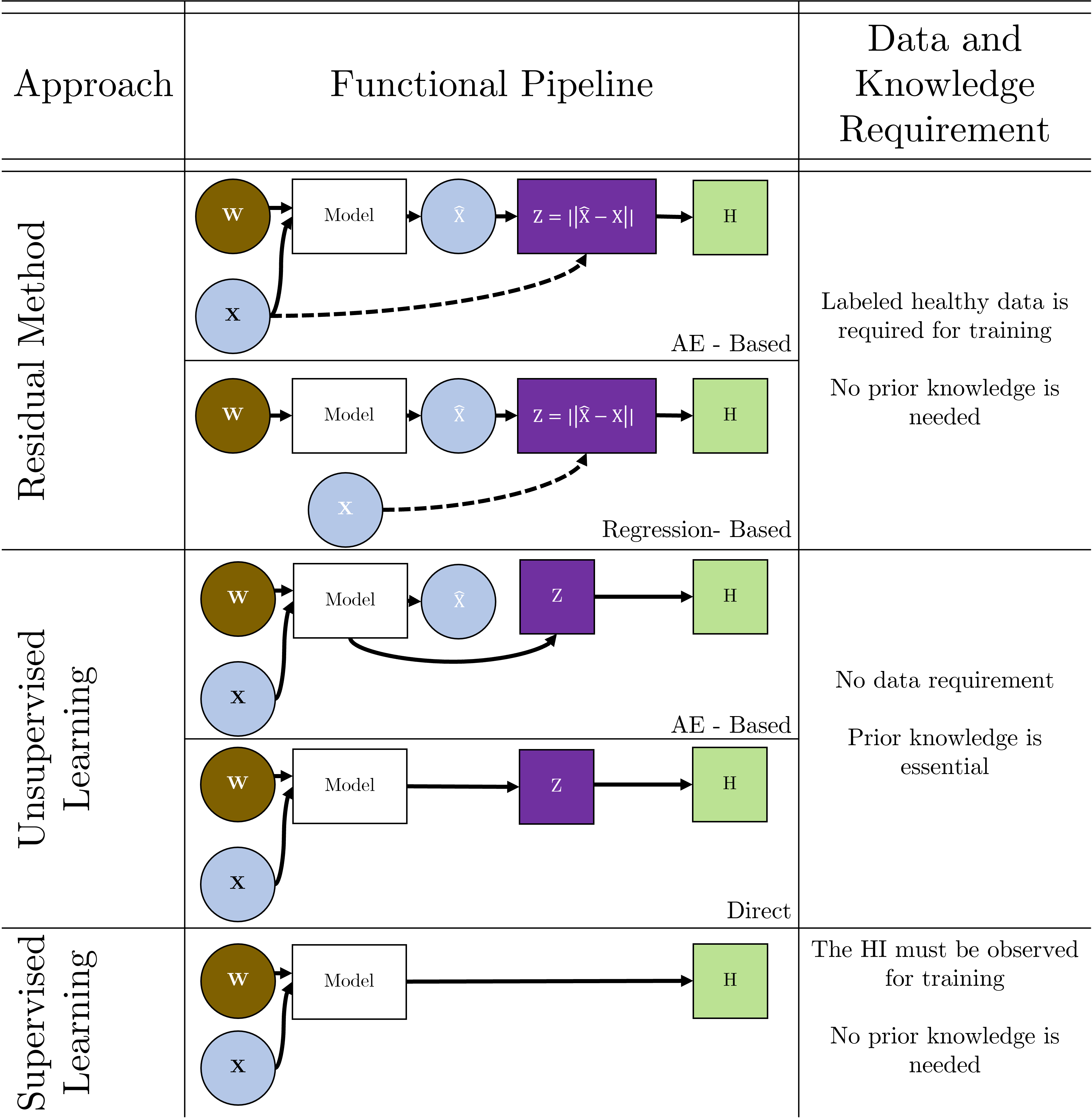}%
    \caption{Overview of different HI estimation methods.}
    \label{fig:Methods_overview}
\end{figure}

In this section, we formally introduce the problem of HI estimation from CM data (Section \ref{sec:back:problem}). We also review related work and discuss three established solution strategies: residual methods (Section \ref{sec:back:semisupervised}), unsupervised (Section \ref{sec:back:unsupervised}), and supervised methods (Section \ref{sec:back:supervised}). These methods serve as a benchmark against which we evaluate the methodology proposed in this work. In Figure \ref{fig:Methods_overview}, we provide an overview of the model functional mappings and data requirements for various HI estimation methods. 


\subsection{Problem Formulation}
\label{sec:back:problem}
We are given multi-variate time series of sensor readings 
\begin{equation}
    X_u = [x_u^1,...,x_u^m]
\end{equation}
of a fleet of N units $(u = 1,...,N)$, each with $m$ observations. Each observation $x_u^i \in \mathbb{R}^p$ is a vector of $p$ raw measurements. We are also given the corresponding scenario–descriptor operating conditions 
\begin{equation}
   W_u = [w_u^1,...,w_u^m]
\end{equation}
for each unit, where each $w_u^i \in \mathbb{R}^k$. The goal is to estimate the state of degradation $Z$ of each unit at each point in time $z_u^i \in \mathbb{R}^z$. The HI of each unit at each point in time $h_u^i$ is then a normalized 1-D representation of the state of degradation $z_u^i$, such that

\begin{equation}
    \{z_u^i \in \mathbb{Z}^z\} \rightarrow \{h_u^i \in \mathbb{R} | 0\leq h_u^i\leq 1 \}
\end{equation}

\subsection{Residual Method HI Estimation}
\label{sec:back:semisupervised}
Safety critical systems undergo comprehensive health monitoring and inspections, allowing the accurate labeling of subsets of CM data as healthy. Alternatively, in certain conditions, as in the case of new units, CM data can be labeled healthy, assuming minimal degradation during the initial operational cycles. Therefore, one of the most common methods for HI estimation in the literature is the residual method \cite{malhotra2016multi,xu2020constructing,she2020sparse, zhai2021enabling,lee2021data,koutroulis2022constructing,zgraggen2022uncertainty,chen2022physics,rao2023speed}.

The residual method uses CM data which is labeled as healthy to train a model $f(\bullet)$ emulating healthy system responses $X$. Once the model is trained, an HI is estimated from the reconstruction residual $r$ of current CM data $X$, which is given by: 
\begin{equation}
    r = |f(\bullet)-X|
\end{equation}

In the final step, the HI is found by reducing the dimensionality of $r$ and normalizing the resulting one-dimensional projection in the range [0, 1]
\begin{equation}
    r \in \mathbb{R}^p \rightarrow h \in \mathbb{R}
\end{equation}

The residual method typically works well under the hypothesis that the training dataset is representative of a healthy system, meaning that small reconstruction errors are typically indicative of healthy inputs, while large reconstruction errors are typically indicative of faulty operations that were not observed during training. For the residual model to function effectively, it is essential to consider variations in operating conditions. This ensures that the model can discern alterations in sensor readings that are unrelated to degradation.

Two types of residual methods are often proposed depending on the inputs of the model $f(\bullet)$. The residual method has been used with an asymmetric AE to reconstruct the sensor readings based on information about the operating conditions and the historical sensor readings \cite{de2023developing,hsu2023comparison}. Another approach is the regression residual method that maps the sensor readings based on the operating conditions \cite{lovberg2021remaining,hsu2023comparison}. The difference between the two methods is visualized in Figure \ref{fig:Methods_overview}. Previous research indicates that the regression-based residual method may be preferable in situations where data collection is limited, but may not perform as well in cases where the sensor readings contain outliers \cite{chalapathy2019deep}.

\subsection{Unsupervised HI Estimation}
\label{sec:back:unsupervised}

In the general scenario of complex systems, the acquisition of labeled data indicating whether or not a system is healthy or the extent of degradation can often prove challenging. In such a scenario, unsupervised learning methods become essential. AE models are a popular unsupervised technique for HI estimation, aiming to learn a representation of unlabelled CM data encompassing both healthy and degraded conditions. For instance, de Beaulieu et al. \cite{de2022unsupervised} showed that, in certain scenarios, AEs can reveal degradation patterns in their latent space in a case study involving turbofan engines. 

Another commonly employed technique is Principal Component Analysis (PCA). Schwartz et al. \cite{schwartz2022unsupervised} achieved successful HI estimation for turbofan engines by leveraging the first principal components extracted from sensor reading data using Kernel PCA. Both PCA and its various extensions have been shown to perform comparably to AEs in certain scenarios.

While the mentioned methods showed success in specific cases using subsets of the CMAPSS turbofan dataset with constant operating parameters, as we reveal in Section \ref{sec:Ablation} of our case study, their effectiveness diminishes in scenarios where operational conditions mask degradation's impact on sensor readings.

\subsection{Unsupervised Hybrid HI Estimation}
Hybrid methods integrating prior knowledge with unsupervised models have emerged as a promising solution strategy for HI estimation \cite{li2024review}. Following the taxonomy of hybrid models proposed in \cite{Laura} and \cite{Karniadakis2021}, these methods can be classified based on \textbf{where prior knowledge} is integrated into the machine learning pipeline and \textbf{what type of knowledge} is integrated. 

Regarding \textbf{where prior knowledge} is integrated, we refer to three main hybridization strategies. Observational bias involves augmenting the training data with synthetic data or derived features that reflect underlying prior knowledge, serving as a weak mechanism to embed knowledge into machine learning models. Inductive bias focuses on crafting specialized model architectures that implicitly incorporate the additional knowledge. Learning bias seeks to infuse prior knowledge by modifying the model's learning algorithm, ensuring that the model simultaneously fits the observed data and approximately adheres to a set of specified constraints. 

Beyond hybridization strategies, \textbf{the type of knowledge} integrated plays a pivotal role in the performance of hybrid methodologies. Two primary sources of knowledge stand out:  system-specific high-fidelity knowledge, such as simulators, while another group leverages general low-fidelity knowledge.

\paragraph{System-Specific Knowledge}
The application of \textit{observational bias} in combination with system-specific knowledge for HI estimation is demonstrated in the research conducted by Magadan et al. \cite{magadan2023robust}. In their work, a model was trained using features that were extracted by considering interesting bearing frequencies, which were identified based on prior knowledge about the system. Another instance of observational bias is seen in the work of Biggio et al. \cite{biggio2023ageing}. The authors used a battery simulator to train a model with multiple degradation parameters. The resulting model was then capable of extracting health information from actual data. 
An \textit{inductive bias} strategy is proposed in the work of Guo et al. \cite{guo2022health}. The authors combine a data-driven HI with an HI based on knowledge about the specific system. A similar methodology for inductive bias can also be found in \cite{jahromi2009approach,aizpurua2019improved}.


\paragraph{General Knowledge}
Despite these advancements, there are still challenges with methods that use system-specific prior knowledge. Namely, specific knowledge may not apply to diverse systems with different degradation patterns. Additionally, relying on specific knowledge such as predefined linear or nonlinear functions for degradation can bias the estimation of unit-specific degradation patterns, limiting the model's ability to capture unique characteristics. 

A few studies have leveraged general knowledge that is not specific to a single system. This prior knowledge is typically based on the expectation of the HI in these complex systems. The most commonly used piece of knowledge is the understanding that the estimated HI should exhibit certain characteristics, such as being monotonic, correlating with the operational cycle time, and having a consistent threshold for failure across a fleet of units. 

Researchers have used the knowledge about the expectation for the HI in two ways. The first is by using it to guide the selection of data features that mirror the HI’s desired characteristics, thereby introducing an observational bias. This method is evident in the works in \cite{bejaoui2020data,wang2020hybrid,li2023feature}. The second method constructs the learning objective function of the model tasked with HI estimation. This introduces a learning bias into the model, steering the learning process towards solutions that are consistent with our prior understanding of the HI. This method is showcased in the research conducted in \cite{liu2013data,song2018statistical,QIN2023101973,chen2021health,li2020shape,wang2020deep, wen2021generalized, wang2023deep}. 

While the use of general knowledge can lead to the development of a more universal model for HI estimation, it’s not always clear whether this general knowledge is adequate to enhance model performance. This uncertainty arises from the limited informativeness of the knowledge employed.

\subsection{Supervised HI Estimation}
\label{sec:back:supervised}
Supervised learning methods can be used for HI estimation when the state of degradation of a system is directly observable. For instance, in controlled laboratory experiments with battery usage, one indication of degradation is declining performance, often reflected in phenomena like capacity fade. Following analytical calculations of capacity fade, it can function as a proxy for HI and be utilized to train supervised models \cite{roman2021machine,ng2020predicting,zhou2023light,fan2020novel}. \edit{Similarly, in the field of fracture mechanics, where observations of crack length serve as a degradation proxy \cite{liu2008new,kunzelmann2023prediction}, and in the field of machining tool wear, where wear can be measured directly \cite{wang2020physics}.}

However, for most complex systems, the degradation of a system is complex, affecting multiple components, and is unobservable without a detailed inspection. In these cases, supervised learning methods are not applicable since no desired output labels exist.

\subsection{Overview of Related Work}
\label{sec:limitations_of_related_work}
\edit{ Our analysis of the related work, summarized in Table \ref{tab:LiteratureReview}, reveals prevailing trends in HI estimation. Labeled data approaches are notably dominant, as seen in the widespread use of residual methods (RM) and supervised learning (SL) techniques. Although these categories typically do not involve hybrid approaches, there are notable exceptions. For example, in \cite{chen2022physics}, the authors employ a residual approach augmented with physics-informed fault signatures for gearbox degradation discovery, introducing an observational bias. Similarly, in the supervised learning approach of \cite{wang2020physics}, the authors integrate a physical model of tool cutting, fitted from observed data, to augment observed tool wear data with synthetic features generated by the model.}

\edit{However, a specific focus on purely hybrid unsupervised methods reveals a significant research gap: the scarcity of models that employ multiple hybridization strategies. As discussed in  \cite{li2024review}, a combined approach could significantly enhance the flexibility and robustness of hybrid methods. Additionally, the table underscores a critical issue of limited generalizability; only the work by Chen et al. \cite{chen2021health} has demonstrated their methodology across multiple case studies. This restriction highlights concerns about the broad applicability of existing methods. For wider utility, HI estimation methods need to be adaptable and effective across various systems.}

\begin{table}[h!]
\centering
\begin{tabular}{ccccl}
\hline \hline
\textbf{Reference}                  & \textbf{Approach} & \begin{tabular}[c]{@{}l@{}}\textbf{Hybrid}\\ \textbf{Strategy}\end{tabular} & \begin{tabular}[c]{@{}l@{}}\textbf{General}\\ \textbf{Knowledge}\end{tabular}   & \textbf{Case Study}              \\ \hline \hline
\cite{malhotra2016multi}               & RM           & -                            & -             & Turbofan (CMAPSS)             \\ \hline
\cite{xu2020constructing}              & RM           & I                            & -             & Bearings                      \\ \hline
\cite{she2020sparse}                   & RM           & -                            & -             & Bearings                      \\ \hline
\cite{zhai2021enabling}                & RM           & -                            & -             & Turbofan (CMAPSS)             \\ \hline
\cite{lee2021data}                     & RM           & -                            & -             & Turbofan (DASHlink)           \\ \hline
\cite{koutroulis2022constructing}      & RM           & I                            & -             & Turbofan (N-CMAPSS)           \\ \hline
\cite{zgraggen2022uncertainty}         & RM           & -                            & -             & Wind Turbine                  \\ \hline
\cite{chen2022physics}                 & RM           & O                            & -             & Gearbox                       \\ \hline
\cite{rao2023speed}                    & RM           & -                            & -             & Gearbox                       \\ \hline
\cite{de2023developing}                & RM           & -                            & -             & Turbofan (N-CMAPSS)           \\ \hline
\cite{hsu2023comparison}               & RM           & -                            & -             & Turbofan (N-CMAPSS)           \\ \hline
\cite{lovberg2021remaining}            & RM           & -                            & -             & Turbofan (N-CMAPSS)           \\ \hline
\cite{de2022unsupervised}              & UL           & -                            & -             & Turbofan (CMAPSS)             \\ \hline
\cite{schwartz2022unsupervised}        & UL           & -                            & -             & Turbofan (CMAPSS)             \\ \hline
\cite{magadan2023robust}               & UL           & O                            & -             & Bearings                      \\ \hline
\cite{biggio2023ageing}                & UL           & O                            & -             & Battery                       \\ \hline
\cite{guo2022health}                   & UL           & I                            & -             & Electric transformer               \\ \hline
\cite{jahromi2009approach}             & UL           & I                            & -             & Electric transformer                \\ \hline
\cite{aizpurua2019improved}            & UL           & I                            & -             & Electric transformer                \\ \hline
\cite{bejaoui2020data}                & UL            & O                           & $\checkmark$             & Bearings                      \\ \hline
\cite{wang2020hybrid}                 & UL            & O                           & $\checkmark$              & Bearings                      \\ \hline
\cite{li2023feature}                  & UL            & O                           & $\checkmark$              & Bearings                      \\ \hline
\cite{liu2013data}                    & UL            & L                           & $\checkmark$                & Turbofan (CMAPSS)             \\ \hline
\cite{song2018statistical}             & UL           & L                            & $\checkmark$               & Turbofan (CMAPSS)             \\ \hline
\cite{QIN2023101973}             & UL           & L                            & $\checkmark$               & Turbofan (CMAPSS)             \\ \hline
\cite{chen2021health}                 & UL            & L                           & $\checkmark$              & Battery + Bearings            \\ \hline
\cite{li2020shape}                    & UL            & L                           & $\checkmark$              & Turbofan (CMAPSS)             \\ \hline
\cite{wang2020deep}                   & UL            & L                           & $\checkmark$              & Turbofan (CMAPSS)             \\ \hline
\cite{wen2021generalized}             & UL            & L                           & $\checkmark$              & Turbofan (CMAPSS)             \\ \hline
\cite{wang2023deep}                   & UL            & L                           & $\checkmark$              & Turbofan (CMAPSS)             \\ \hline
\cite{roman2021machine}               & SL            & -                           & -              & Battery                       \\ \hline
\cite{zhou2023light}                  & SL            & -                           & -              & Battery                       \\ \hline
\cite{fan2020novel}                   & SL            & -                           & -              & Battery                       \\ \hline
\cite{liu2008new}                     & SL            & -                           & -              & Materials                     \\ \hline
\cite{kunzelmann2023prediction}       & SL            & -                           & -              & Materials                     \\ \hline
\cite{wang2020physics}                & SL            & O+L                         & -              & Materials                      \\ \hline
\begin{tabular}[c]{@{}l@{}}Proposed\\ Method\end{tabular}                      & UL            & I+L                         & $\checkmark$   & \begin{tabular}[c]{@{}l@{}}Turbofan (N-CMAPSS)\\ \ +Battery\end{tabular}  \\ 
\end{tabular}
\caption{Overview of recent works on HI estimation. The column explanations are as follows: \textit{Approach} indicates HI estimation method (RM - residual method, UL - unsupervised learning, SL - supervised learning), \textit{Hybrid Strategy} shows which hybridization strategy was used (O - observational bias, I - inductive bias, L - learning bias, X - none), \textit{General Knowledge} denotes whether general knowledge was used for a hybrid model ($\checkmark$) or not (X), and \textit{Case study} specifies the systems on which the method was tested.}
\label{tab:LiteratureReview}
\end{table}

\clearpage
\section{Methodology}
\label{Methodology}

\begin{figure}[h!]
    \centering
    \includegraphics[scale = 0.9]{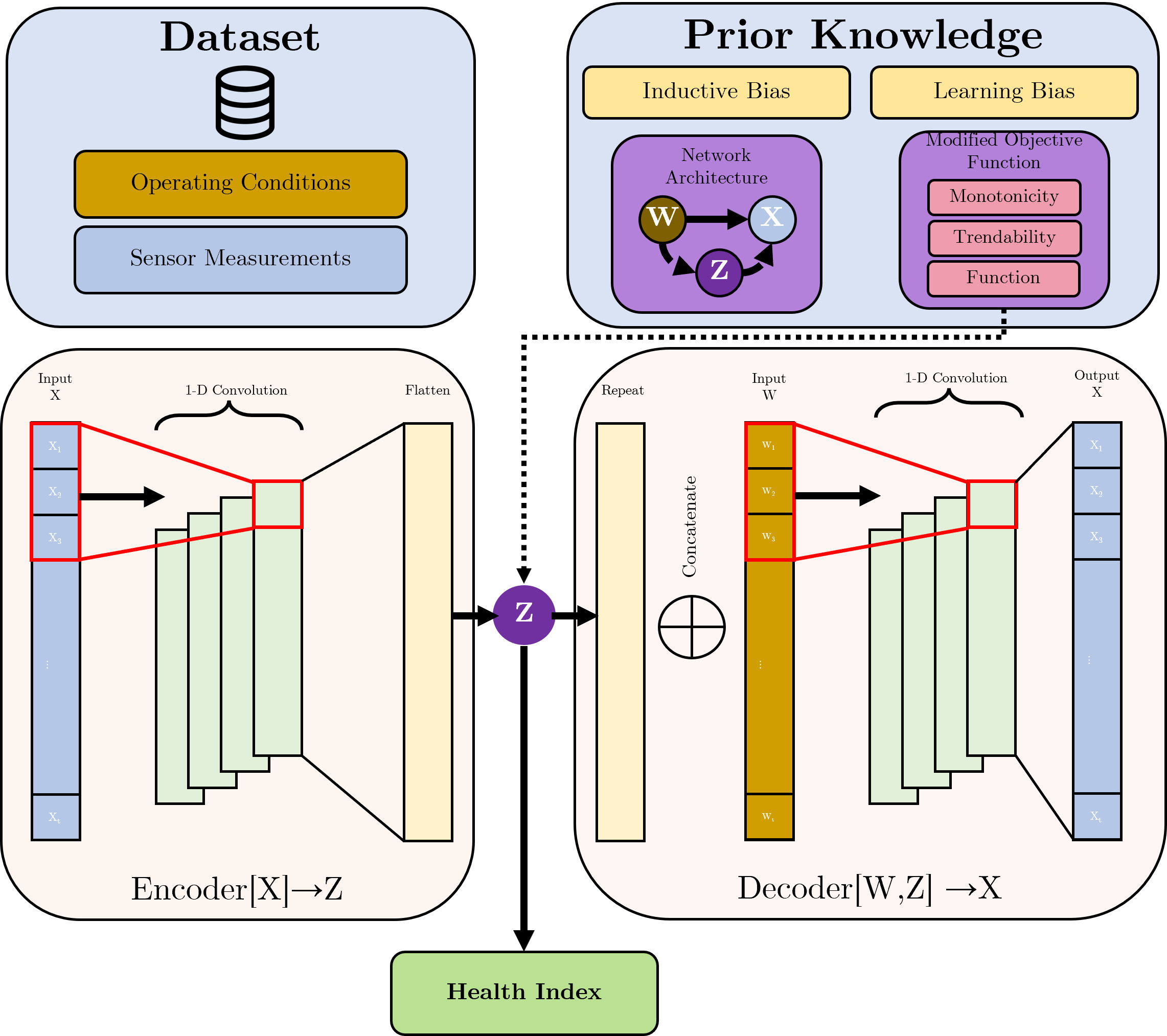}
    \caption{Overview of the proposed unsupervised hybrid method. The proposed method utilizes an AutoEncoder whose architecture is informed by prior knowledge regarding the structure of the HI estimation problem. The encoder processes sensor readings (X) to estimate degradation (Z), while the decoder reconstructs (X) using operating conditions (W) and estimated degradation (Z). The loss function incorporates an additional constraint derived from prior knowledge about degradation's temporal evolution. Finally, the HI is derived from degradation through subsequent post-processing steps.}
    \label{fig:Overview}
\end{figure}

\edit{
To address the challenge of HI estimation from CM data of various complex systems, we propose a novel unsupervised hybrid method based on general knowledge about degradation. To compensate for the potential lack of information of such general knowledge, we propose combining multiple hybridization strategies. Specifically, the method incorporates two key aspects visualized in Figure \ref{fig:Overview}. We introduce an inductive bias directly into the model architecture of the method. This is achieved by utilizing an AutoEncoder whose structure is informed by the causal structure between variables involved in HI estimation. In addition, we apply a learning bias to modify the objective function used to train the model, reflecting expected degradation dynamics in complex systems. In the following section, we present more details about the inductive bias in Section \ref{sec:meth:inductive} and the learning bias in Section \ref{sec:meth:learning}.}

\subsection{Inductive Bias: Derived Model Architecture}
\label{sec:meth:inductive}
\begin{figure}[h!]
\centering
\includegraphics[scale=.7]{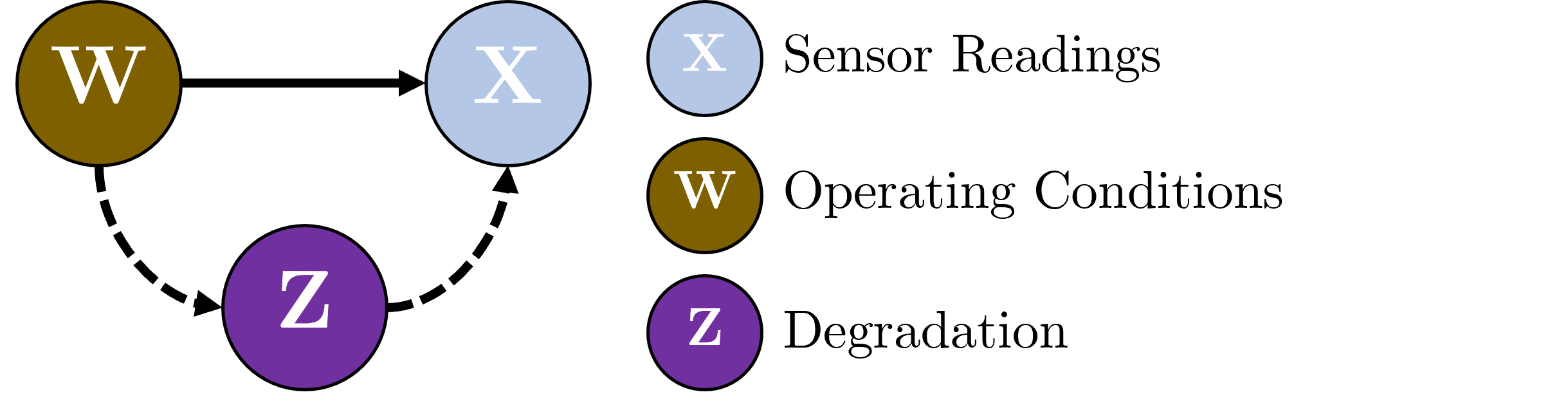}
\caption{Causal representation of variables in a degrading system. $X$ denote the sensor readings, $W$ the operating conditions and $Z$ the degradation. Observable variables are depicted by solid lines, while dashed lines illustrate hidden variables. }
\label{fig:Causality}
\end{figure}

\edit{
In the context of PHM, it is widely accepted \cite{si2015adaptive,nouri2022assessment,alves18group} that the performance of systems, as expressed in the sensor readings ($X$), is typically influenced by both operating conditions ($W$) and by degradation ($Z$). In this study, we show this empirically by leveraging elements from causal theory. Concretely, we use the additive noise model (ANM) \cite{hoyer2008nonlinear} to establish the causal relationships between the variables of interest in a directed acyclic graph (DAG) \cite{pearl2000models} (see Figure \ref{fig:Causality}). The empirical study is presented in more detail in \ref{sec:ANM}.}

\edit{
The causal direction $W \rightarrow X$ can be justified by observing that sensor readings ($X$) vary significantly under different operational conditions ($W$). For example, a commercial aircraft will go through a series of flight stages (e.g., taxiing, take-off, cruise, descend) that affect its sensor recordings. This effect is usually easily observed or recognized. The causal direction $W \rightarrow Z$ represents the influence that operational conditions have on degradation. The stresses and conditions to which a system is subject over its lifetime will have a long-term impact on the system's degradation. 
The third causal direction $Z \rightarrow X$ is a central assumption in prognostics: sensor readings ($X$), which reflect the performance of a system, are subject to changes due to degradation ($Z$), even though this effect may not be as pronounced as the influence of operating conditions ($W$). Formally, we can express the previous causal graph as a structural causal model (SCM) with assignments given as follows:}
\edit{
\begin{align}
  \mbox{Operational conditions } & W := f_1(\epsilon_1) \label{eq1} \\
  \mbox{Degradation }  & Z := f_2(W, \epsilon_2)  \label{eq2}\\
   \mbox{Sensor readings }&  X := f_3(W,Z,\epsilon_3) \label{eq3}
\end{align}
}
\edit{
where $\epsilon_1, \epsilon_2, \epsilon_3$ are jointly independent noise variables and $f_1,f_2,f_3$ are deterministic causal functions. It is important to recognize that these assignments (operator $:=$) are unidirectional, with causal variables on the right and dependent variables on the left.}

\edit{
The SCM implies that operational conditions ($W$) are an independent process. Degradation ($Z$) is dependent and caused by the operational conditions ($W$). The sensor readings ($X$) are caused by both operational conditions ($W$) and degradation ($Z$). As described previously, this structure can be derived empirically from observational data using Algorithm \ref{ANM} in \ref{sec:ANM}.}

\edit{
Understanding the causal relationships between the variables $W$, $Z$, and $X$ is crucial to our work because it enables the development of a more appropriate unsupervised learning architecture \cite{peters2017elements}. Even though it may appear intuitive from Assignment \ref{eq2} to estimate the degradation $Z$ from the operational conditions $W$, this approach is misleading. Formally, the principle of causal conditional independence dictates that knowing the distribution of a cause ($W$) does not provide additional insights into how $W$ influences the effect ($Z$). In simpler terms, even with extensive data about $W$, we cannot directly infer how $W$ affects $Z$ because causality flows in one direction.}

\edit{
On the other hand, the scenario becomes more nuanced when considering "anticausal" learning. This is, if the variable previously considered as an effect ($Z$) becomes a cause for another variable ($X$), then information about the latter ($X$) can be informative about the former ($Z$). In essence, by studying the effect of $Z$ on $X$, we gain indirect insights into the nature of $Z$ itself. Note, however, that $f_3(W, Z)$  captures the combined influence of both $Z$ and $W$ on $X$. Simply estimating $Z$ from $X$ would inherit the confounding effect of $W$, making it difficult to isolate the true effect of $Z$ on $X$.}

\edit{
To address this, we propose a specific model architecture for isolating $Z$. This architecture leverages an AE with an encoder $\mathcal{G}_{\theta}$ and decoder $\mathcal{F}_{\phi}$. The encoder is trained to encode $X$ to an estimated representation of $Z$, denoted as $\hat{Z}$.}

\edit{
\begin{equation}
\mathcal{G}_{\theta}(X) \rightarrow \hat{Z}
\end{equation}
Which is the anticausal direction permitted by the SCM structure. Thereafter, the decoder $\mathcal{F}_{\phi}$ uses $W$ and $\hat{Z}$ to reconstruct $X$.
\begin{equation}
\mathcal{F}_{\phi}(W,\hat{Z}) \rightarrow \hat{X}
\end{equation}
This is consistent with the expression presented in Assignment \ref{eq3}. The proposed AE model is given by:
\begin{equation}
\mathcal{F}_{\phi}(W,\mathcal{G}_{\theta}(X)) = \hat{X}
\end{equation}
It is trained with the following objective function 
\begin{equation}
    L_{MAE} = \frac{1}{m} \sum| X_i - \hat{X_i} |
\end{equation}
}
\edit{
The decoder effectively captures changes in $X$ attributed to variations in $W$ and $Z$. Because the encoder solely relies on $X$ to derive $Z$, the network is compelled to learn crucial information unrelated to $W$. In the ablation study detailed in Section \ref{sec:Ablation}, we show that our proposed architecture performs better than a model that inputs operating conditions into both the encoder and decoder.}



\subsection{Learning Bias: Embedding Cycle Information}
\label{sec:meth:learning}

Important degradation mechanisms in complex systems are typically dominated by operation time. For instance, in turbofan engines, degradation mechanisms such as friction, erosion, and fouling of rotating components, are dominated by cycle operation. Similarly, degradation mechanisms in batteries, such as solid–electrolyte interphase layer growth, lithium plating, or particle fracture, are also dominated by cycling \cite{edge2021lithium}. Therefore, in this subsection, we present how general knowledge about the temporal dependence of the degradation can be embedded into the data-driven pipeline as a learning bias modifying the loss function of the proposed AE with a soft constraint.

We show three potential methods to incorporate the influence of operational cycles on degradation in the architecture: 1) trendability, 2) negative gradient, and 3) HI function derived from reliability theory.  Each method consists of a different soft constraint that is imposed on the latent space of the AE architecture. Notably, these hybridization techniques are referred to as inductive bias because the soft constraints guide the latent space, enabling it to unveil degradation without overly restricting the AE's functionality. The soft constraints are implemented as an additional term in the objective function of the model with the parameter $\lambda$ used to control the significance of the constraint and mitigate the risk of overfitting.

\paragraph{Trendability}
The first method involves imposing a restriction based on the Spearman correlation between $t$ and $Z$. The motivation for imposing a correlation between operation cycles and degradation lies in the necessity to establish a relationship between the increasing age of the equipment and the decreasing HI. The constraint is defined as follows:

\begin{equation}
    L_{C} = \frac{\sum (t_i-\overline{t})(Z_i-\overline{Z})}{\sqrt{\sum (t_i-\overline{t})^2 \sum (Z_i-\overline{Z})^2}}
\end{equation}
Our proposed method is trained with the following objective function:
\begin{equation}
    L = L_{MAE} + \lambda L_{C}
\end{equation}

\paragraph{Monotonicity}
The second method is similar to the first one, but instead, we aim to constrain the gradient of $Z$ with respect to $t$ to be negative, that is $\frac{\partial Z}{\partial t} \leq 0$. The constraint is motivated by the understanding that the rate of $Z$ change concerning $t$ must be negative to derive a monotonically decreasing HI. Compared to the previously mentioned correlation constraint, the negative gradient constraint is more flexible because it does not implicitly impose a linear constraint.
We define the negative gradient constraint as:
\begin{equation}
    L_{NG} = \max\{0,\frac{\partial Z}{\partial t}\}
\end{equation}
Our proposed method is trained with the following objective function:
\begin{equation}
    L = L_{MAE} + \lambda L_{NG}
\end{equation}

\paragraph{HI function derived from reliability theory}
The third method can be used if one knows a function of the expected health index for a given cycle, i.e. $h=g(t)$. The function $g(t)$ might be known from prior knowledge or could be derived from the history of HI. We propose a method to derive $g(t)$ from the history of HIs inspired by reliability theory. For more information see \ref{Weibull}. Compared to the two previous constraints, the functional constraint is the most restrictive and system-specific, since it is individual to the specific system being investigated. We define a function $g(t)$ given by:

\begin{equation}
    g(t)= C - ((t(log(1 - P))^{-1/\beta})A)^B
\end{equation}
Where the parameters A, B, C, $\beta$ are estimated from data. The constraint is then given by:
\begin{equation}
    L_{F} = \frac{1}{m} \sum| g(t_i) - Z_i |
\end{equation}
Our proposed method is trained with the following objective function:
\begin{equation}
    L = L_{MAE} + \lambda L_{F}
\end{equation}

\section{Case Studies}
\label{Case Studies}
The proposed method is demonstrated and evaluated in two case studies featuring distinct complex systems. These case studies vary in multiple aspects, including the obvious difference between an eletrochemical process and a thermo-kinetic process. but also the rate of degradation with respect to operational cycles and the manifestation of the impact of degradation.  By examining these diverse scenarios, we aim to illustrate the robustness and applicability of our proposed method across multiple complex systems.

Section \ref{sec:casestudy:turbofan} presents the airplane turbofan engine case study, and Section \ref{sec:casestudy:battery} introduces the lithium battery case study. In Section \ref{sec:casestudy:preprocessing}, we describe the preprocessing of the datasets. Section \ref{sec:casestudy:hi} describes how the HI is extracted for the considered methods. The implemented network architectures are briefly presented in Section \ref{sec:results:networks}. Finally, we discuss how to evaluate HI estimation methods in Section \ref{sec:meth:eval}.

\subsection{Dataset: Aircraft Turbofan Engines}
\label{sec:casestudy:turbofan}

\begin{figure}[h!]%
    \centering
    \subfloat[\centering Flight altitude conditions]{{\includegraphics[scale = 0.8]{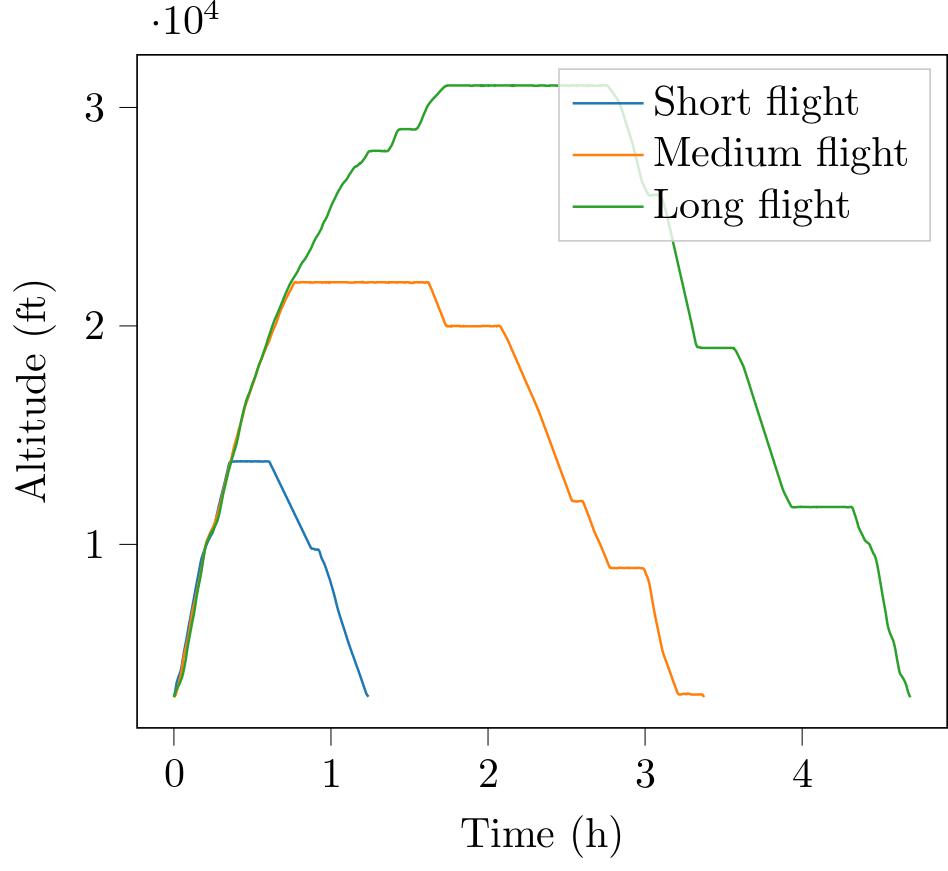} }}%
    \subfloat[\centering Health Index.]{{\includegraphics[scale = 0.8]{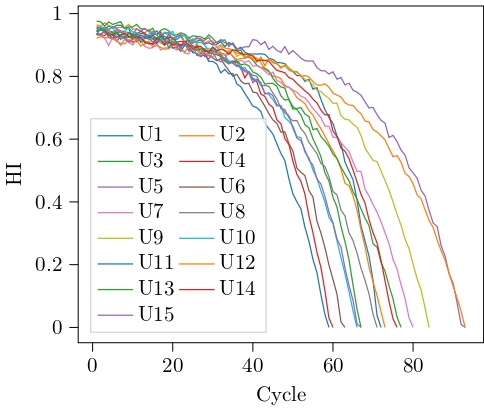} }}%
    \caption{Typical flight conditions and HI for units in the N-CMAPSS dataset.}
    \label{fig:cmapss-hi}
\end{figure}

The new Commercial Modular Aero-Propulsion System Simulation (N-CMAPSS) dataset \cite{arias2021aircraft} provides full run-to-failure degradation trajectories of turbofan engines. From the eight available data subsets within the N-CMAPSS dataset, we consider the set DS003, which is characterized by a failure mode that affects the low-pressure turbine efficiency and flows in combination with the high-pressure turbine efficiency.  Each unit within the fleet contains 14 observable sensor measurements, denoted as $X$, which are recorded from an initial health condition until engine failure (run-to-failure data). In addition to the sensor measurements, the corresponding operating conditions $W$ are available. The operating conditions include altitude, Mach number, throttle-resolved angle, and total temperature at the fan inlet. The units are divided into three flight classes depending on whether the unit is operating short-length flights, medium-length flights, or long-length flights. Figure \ref{fig:cmapss-hi}(a) displays the typical altitude conditions for randomly selected flight cycles and units.

The N-CMAPSS dataset models degradation at the component level through initial, normal, and abnormal degradation stages. As a result of degradation, a HI is computed in the form of a non-linear mapping of multiple operational margins taken at reference conditions. The resulting HI was used to declare system failure when its value reached 0. The dataset also incorporates between-flight maintenance by permitting improvements in engine health parameters within allowable limits. This ground truth HI (i.e., $h_{GT}$) is available for DS03, and will be used for evaluation purposes only. Figure \ref{fig:cmapss-hi}(b) illustrates the ground truth Health Index (HI) for the fleet of units.

\subsection{Dataset: Li-ion Batteries}
\label{sec:casestudy:battery}

The randomized battery usage dataset from the NASA Ames Prognostics Center of Excellence repository \cite{bole2014adaptation} was considered to further evaluate the proposed methodology. The dataset encompasses data from individual 18650 LCO cells, each subjected to cycles of charging and discharging under randomized protocols.

The most common physical aging mechanisms observed in batteries are graphite exfoliation, loss of electrolyte, solid electrolyte interface layer formation, continuous thickening, lithium plating, etc. \cite{sui2021review}. Consequently, the battery aging process gives rise to two primary changes to the battery electrodynamics, which impact its performance: capacity fade and increase in internal resistance. In this particular case study, we will focus on capacity fade as the selected HI for the batteries under examination.

\begin{figure*}[h!]%
    \centering
    \subfloat[\centering Random discharge cycles.]{{\includegraphics[scale = 0.30]{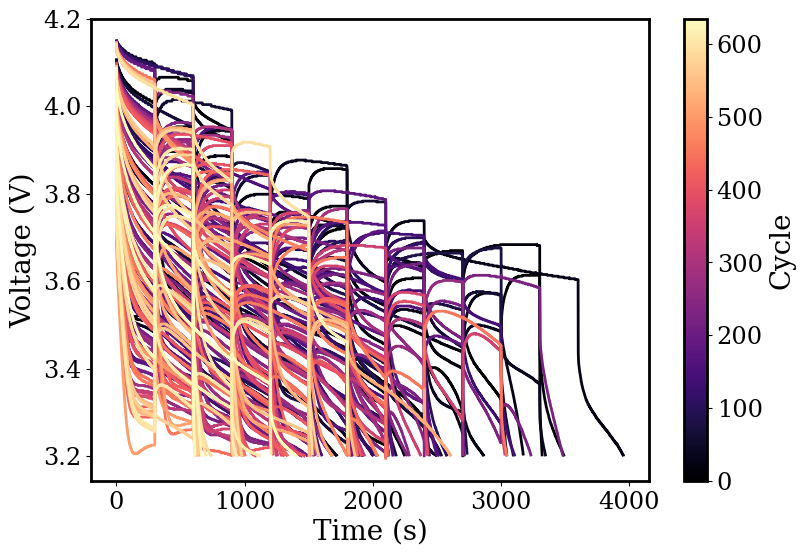}}}%
    \subfloat[\centering Reference discharge cycles.]{{\includegraphics[scale = 0.30]{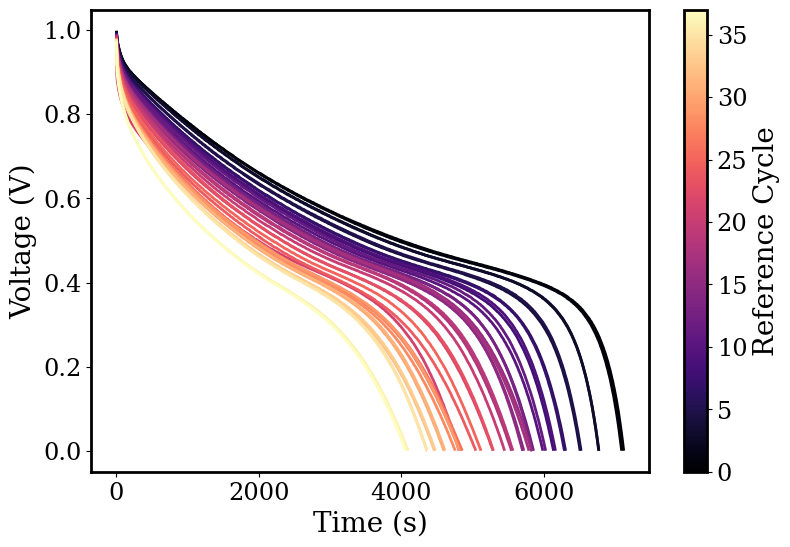}}}%
    
    \subfloat[\centering Calculated capacity.]{{\includegraphics[scale= 0.30]{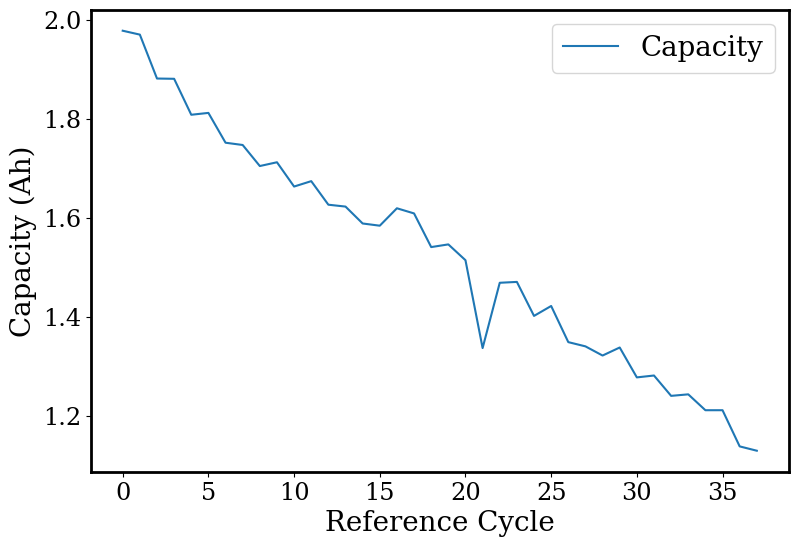} }}%
    \caption{Example of random discharge curves available for model training, and reference discharge curves used to calculate capacity values and establishing ground truth HI.}
    \label{fig:battery_hi}%
\end{figure*}

To simulate real-world battery usage scenarios, we only consider the randomized battery discharge cycle data. For each battery in the dataset, voltage and temperature measurements (X) were recorded during different operational conditions (W), defined by the applied current during the discharge process. Random discharge cycles are illustrated in Figure \ref{fig:battery_hi}(a).

To estimate the ground truth HI, we analytically calculate capacity values from reference discharge cycles with constant current. The current capacity $Q$ of the battery is calculated as $Q = \frac{1}{3600} \int W(t)\,dt$. Illustrated in Figure \ref{fig:battery_hi}(b) are the constant current reference discharge cycles, and in Figure \ref{fig:battery_hi}(c) the calculated capacity values for each corresponding cycle. 

In battery-related contexts, it is more common to use the notation of State of Health (SOH) instead of the HI. The SOH is defined as the ratio between the present capacity and the nominal capacity ($SOH=Q/Q_{nominal}$). In this paper, a failure of a battery will be defined once SOH is less than 60\%. Since HI and SOH have similar meanings, the terms will be used interchangeably in this paper.

\subsection{Pre-Processing}
\label{sec:casestudy:preprocessing}
In all our experiments, we initially performed min-max normalization on both $X$ and $W$ to scale their values within the range [0,1]. To improve computational efficiency, we also reduced the data sampling frequency. Specifically, for the turbofan dataset, we decreased the frequency from 1Hz to 0.1Hz and from 1-Hz to 0.5-Hz for the battery dataset. 

Following these pre-processing steps, the next stage involved segmenting the data into fixed-length windows of size $S$. In the context of optimizing both the residual and supervised methods, we conducted a grid search to identify the optimal window size for each. Consequently, for the residual method and the supervised model, we employed a sliding window with a length of $S = 50$ for the turbofan dataset and $S = 200$ for the battery dataset. In contrast, our proposed method involved windowing the data based on individual operational cycles. Since operational cycles can vary in duration, we employed padding with zeros at the end of each cycle to equalize their lengths. The window size was determined by selecting the minimum integer that aligned with the length of the longest cycle. For the turbofan dataset $S = 2030$, and the battery dataset $S = 2160$.

Windowing whole operational cycles is the preferred methodology since it makes post-processing of the HI straightforward. For the residual method and supervised model windowing whole cycles was not possible since it created unwanted artifacts due to the zero-padding. It is also worth noting that our proposed method also works with shorter window sizes, but as previously mentioned due to the nature of easier HI construction, we have opted to window whole cycles. This will be discussed in more detail in Section \ref{sec:casestudy:hi}.

\subsection{Constructing the HI}
\label{sec:casestudy:hi}
In the context of our research, the extraction of HIs depends on the chosen methodology and the specific case study at hand. In the supervised model, the HI is directly estimated from the model's output. In contrast, the residual method requires a multi-step process. We first compute the residual error for each prediction window. Next, we flatten each window and employ the Principal Component dimensionality reduction technique, effectively transforming the residual vector, denoted as $r \in \mathbb{R}^{p}$, into a $\mathbb{R}$.  Since the supervised and the residual methods use short window sizes, which also have overlapping observations, we also average the resulting HI per cycle. 

Meanwhile, our proposed method offers a more streamlined HI extraction process. Degradation is extracted directly from the output of the encoder, i.e., the latent layer corresponding to $Z$ in Figure \ref{fig:Overview}. Since whole operational cycles are used as input, there is no need to smooth the resulting HI. 

For the turbofan dataset, we normalize the resulting HI to be within $[0,1]$. For the battery dataset, we normalize the resulting HI to be within $[0\%,100\%]$. As mentioned earlier, the resulting HI for the battery dataset corresponds to SOH, a more frequently used metric for expressing the health of the battery in literature.

\subsection{Network Architectures}
\label{sec:results:networks}
In this section, the network architectures of the considered HI estimation methods are described.

\textbf{Residual method - AE}.   The asymmetric-AE residual model is shown in Figure \ref{fig:Methods_overview}(a). The model is trained to reconstruct X when W and X are used as input. The architecture of the asymmetric-AE residual model used here comprises four identical 1-D CNN layers with 64 filters of size 11 and with ReLU activation functions. 

\textbf{Residual method - Regression}. The regression type residual model implemented in this study is shown in Figure \ref{fig:Methods_overview}(b). We train a model to predict X given W as input. The model contains four identical 1-D CNN layers with 64 filters of size 11 and with ReLU activation functions. 

\textbf{Proposed method - Unsupervised Hybrid AE}.
The structure of the proposed model is shown in Figure \ref{fig:Overview}. The model is composed of two parts: an encoder and a decoder. Both of these parts are built using 1D-convolution layers. Specifically, the encoder takes as input X and passes it through three 1-D convolution layers with a number of filters equal to [128,64,16]. Afterward, the output of the encoder is flattened and passed through a fully connected layer with one neuron. The output of the fully connected layer is Z. The decoder concatenates Z and W as input and passes it through three 1-D CNN layers with a number of filters equal to [16,64,128]. The last layer of the decoder is a fully connected layer with a number of neurons equal to the dimensionality of X. The loss function of the model is modified based on the chosen general knowledge. We do not perform any fine-tuning of the constraint parameter $\lambda$ and set it to 1. 

\textbf{Supervised model.}
We train a supervised model for HI estimation inspired by the architecture of \cite{fan2020novel}. The model uses as input X and W to predict the HI in a supervised manner. The model contains four identical 1-D convolution layers with the number of filters set to 64 and a kernel width of 11. Each convolution layer is followed by batch normalization and a max pool layer of size 2. The output of the final convolution layer is flattened and then passed through a dense layer with the output size of 1, corresponding to the HI.

The optimization of the network’s weights is carried out with mini-batch stochastic gradient descent (SGD) and with the Adam algorithm. The training hyper-parameters for each model are given in Table \ref{tab:hyper}.

\begin{table}[h!]
\centering
\begin{tabular}{llllll}
\hline \hline
Dataset                  & Model      & Window size & Epochs & Batch size & Learning rate \\ \hline \hline
CMPAPSS \rule{0pt}{1.0\normalbaselineskip} &                  Supervised & 50         & 20     & 512        & 1e-4          \\
                         & Residual   & 50         & 20     & 512        & 1e-4          \\
                         & Proposal   & 2030        & 20    & 20         & 1e-4          \\ \hline
Battery \rule{0pt}{1.0\normalbaselineskip} &                  Supervised & 200         & 20    & 1024       & 1e-4          \\
                         & Residual   & 200         & 20    & 1024       & 1e-4          \\
                         & Proposal   & 2160        & 20    & 128        & 1e-4         
\end{tabular}
\caption{Hyperparameters of investigated methods.}
\label{tab:hyper}
\end{table}

\subsection{Evaluation}
\label{sec:meth:eval}
\begin{figure}[h!]
    \centering
    \includegraphics[scale = 0.7]{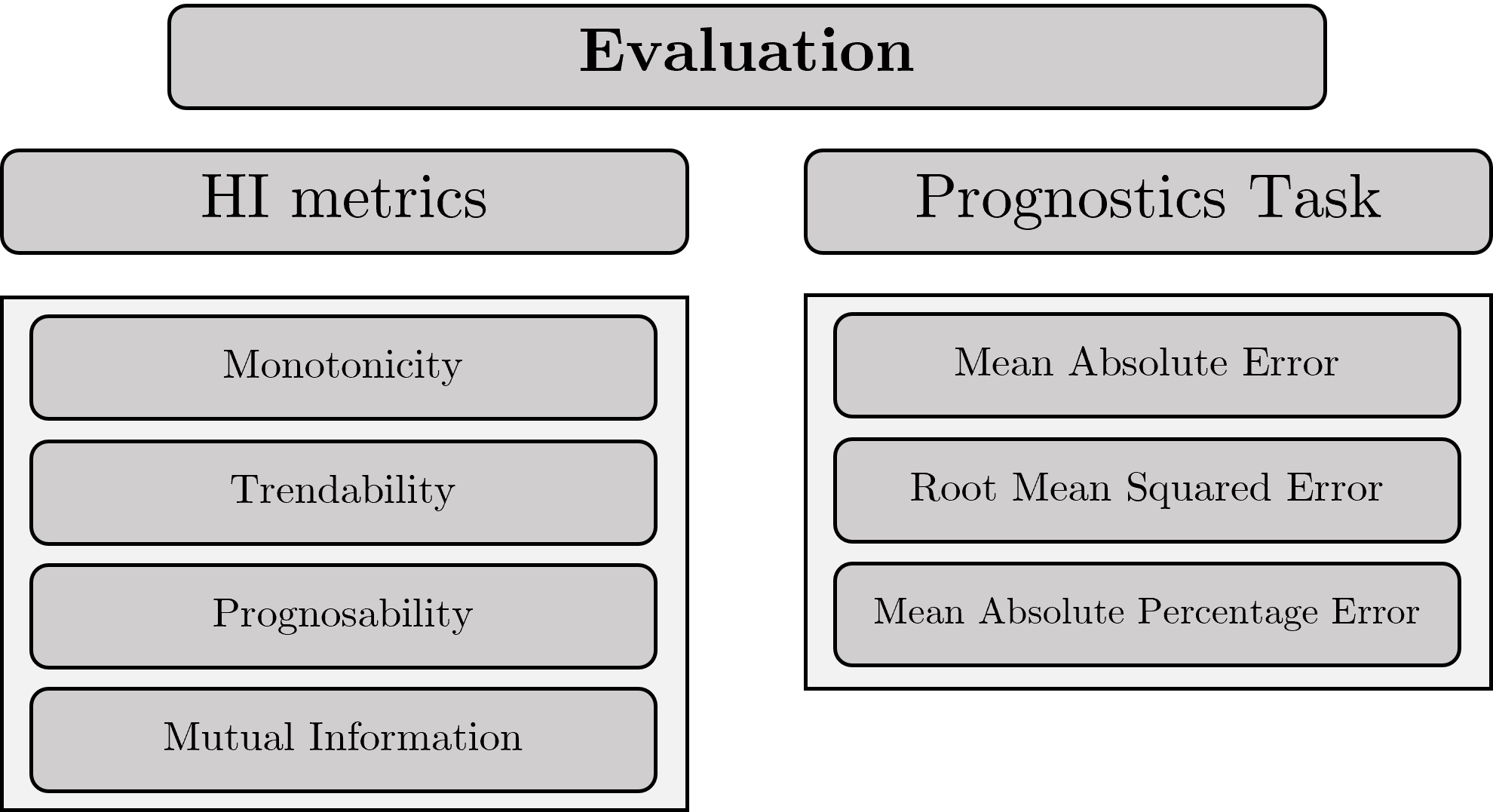}
    \caption{Evaluation methodology overview}
    \label{fig:Evaluation_Overview}
\end{figure}

Based on the state-of-the-art HI evaluation methodology \cite{nguyen2021automated}, we compare and analyze the performance of the proposed method for HI estimation based on two evaluation aspects: quality of the HI and impact on prognostic performance when the HI is used for RUL prediction task. For each of the two aspects, we consider evaluation metrics that are defined in the following sections. An overview of the evaluation methodology is given in Figure \ref{fig:Evaluation_Overview}.

\subsubsection{HI Criteria}
There are several desirable properties that an HI should exhibit to represent the degradation of a system accurately. Although initial health conditions and operational modes can cause some variability in the estimated HIs, it is still desirable for them to demonstrate consistent behavior.

In this work, we employ the following criteria for HI evaluation:
\begin{itemize}

\item  \textbf{Monotonicity (Mon)} measures the tendency for the HI to consistently increase or decrease \cite{coble2010merging,mao2021construction}.  
    
\item  \textbf{Trendability (Tren)} is used to evaluate the degree to which the HIs of a fleet of systems have a similar shape and underlying form
\cite{coble2010merging,mao2021construction}.

\item  \textbf{Prognosability (Prog)} is used to evaluate consistent HI behavior towards the end of life of units  \cite{coble2010merging,mao2021construction}.

\item  \textbf{Mutual Information (MutInf)} score quantifies the information obtained about RUL by observing HI \cite{nguyen2021automated}.  
\end{itemize}
For more information about each of the HI criteria see \ref{HI_criteria}.

\subsubsection{Prognostic Performance}
A key objective of HI estimation is to enhance the performance of prognostic models. To validate the effectiveness of the proposed HI estimation techniques, a baseline prognostic model is needed. The sensor signals, operating conditions, and cycles are used as inputs to predict RUL. The model is given by:
\begin{equation}
    G(X,W,t)=RUL
\end{equation}
To test whether the estimated HIs increase prognostic performance, HIs are used to augment the input space.
\begin{equation}
    G(X,W,t,h)=RUL
\end{equation}

The chosen RUL model for both case studies is based on a 1D-CNN architecture, as outlined in the work by Chao et al. \cite{chao2022fusing}. The parameters of the model are kept fixed for all experiments. 

In particular, for the turbofan dataset, the CNN architecture includes five layers. Three initial convolutional layers utilize filters of size 10. The first two convolutions have ten channels, and the last convolution has only one channel. Zero padding is applied to maintain the feature map throughout the network. The resulting 2D feature map is flattened, leading to a 50-way fully connected layer followed by a linear output neuron. ReLu serves as the activation function for the network. The window length matches that of the residual and supervised HI estimation models (s=50). Similarly, the battery dataset adopts a comparable CNN architecture to the turbofan dataset. The distinction lies in all three convolution layers having ten channels, and the last fully-connected layer has a size of 200 (matching the window length).

Prognostic assessments are commonly based on metrics such as mean absolute error (MAE), root mean squared error (RMSE), and mean absolute percentage error (MAPE). Furthermore, we compute '\% Average Improvement' which denotes the performance improvement over the baseline model upon inclusion of HI, with the same consideration being given to MAE, RMSE, and MAPE.

\section{Results}
\label{Results}
In this section, we analyze the performance of the proposed model on two case studies: the turbofan dataset and the battery dataset. We evaluate a situation where the training and test datasets are of the same distribution in Section \ref{Results:InDistribution}. Furthermore, we investigate HI estimation performance in the context of out-of-distribution testing in Section \ref{Results:Abalation}. The resulting structure of this section is shown in Table \ref{tab:Results_summary}.

\begin{table}[h!]
\centering
\begin{tabular}{lll}
\hline \hline
Experiment                    & Turbofan & Battery \\
\hline \hline
In Distribution \rule{0pt}{1.0\normalbaselineskip}      &  Section \ref{Results:Turbofan}        & Section \ref{Results:Battery}         \\
Out of Distribution &  Section \ref{Results:Turbofan_OOD}        & Section \ref{Results:Battery_OOD}       
\end{tabular}
\caption{Overview of the results section.}
\label{tab:Results_summary}
\end{table}

\subsection{In-Distribution Testing}
\label{Results:InDistribution}

The initial set of experiments will focus on the in-distribution case, where the term "in-distribution" refers to training and testing data originating from similar data distribution. The N-CMAPSS dataset encompasses engines classified into three distinct flight classes. The operational conditions of engines are influenced by their assigned flight class, subsequently impacting degradation patterns. 
To create a balanced dataset for training and testing we use the data split given in Table \ref{table:dataset}.

\begin{table}[h!]
\centering
\begin{tabular}{l|l|l}
 \hline \hline
Flight Class \rule{0pt}{1.0\normalbaselineskip} & Training Units & Testing Units \\
 \hline \hline 
Short      \rule{0pt}{1.0\normalbaselineskip}      & U1, U5, U9                & U12, U14 \\
Medium                                             & U2, U3, U4, U7                & U15 \\
Long                                               & U6, U8                 & U10, U11, U13
\end{tabular}
\caption{In-distribution training set-up for N-CMAPSS dataset.}
\label{table:dataset}
\end{table}

Similarly, for the battery dataset, batteries undergo discharging with randomized loads selected from a given distribution. These batteries are categorized into three load classes based on the selected load distribution: uniform, skewed high, and skewed low. The load class significantly influences the operational conditions of the batteries, consequently affecting degradation patterns. We utilize four subsets for training and testing our proposed method. Table \ref{table:RWBattery} provides a summary of the data subsets, along with the division into training and test sets.

\begin{table}[!h]
\centering
\begin{tabular}{l|l|l}
 \hline \hline
Load profile \rule{0pt}{1.0\normalbaselineskip}   & Training Batteries & Testing Batteries \\
 \hline \hline 
Uniform     \rule{0pt}{1.0\normalbaselineskip}       & RW4, RW5 & RW6 \\
Uniform   &   RW1, RW7              & RW8  \\
Skewed High  & RW17, RW19 &  RW20 \\
Skewed Low & RW13, RW14, RW15&  RW16
\end{tabular}
\caption[Caption LOF]{In-distribution training set-up for NASA battery dataset.\protect\footnotemark}
\label{table:RWBattery}
\end{table}

\footnotetext{Certain subsets of batteries were omitted from consideration: one due to a complex charging process, two because experiments were conducted at 40°C external temperature while the rest were at room temperature, and three (RW2, RW3, and RW18) due to corrupted temperature readings.}

\subsubsection{In-Distribution Testing - Turbofan}
\label{Results:Turbofan}
The HI metrics obtained from the six evaluated models are shown in Table \ref{table:ResultsHI_turbofan}. We also report the mean absolute percentage error (MAPE) between the estimated HI and the ground truth HI.

Comparing residual-based methods, both trained with CM data from the initial 20 flight cycles of each training unit, we observe that the regression-based residual method ($h_{r}^b$) outperforms the AE-based method ($h_{r}^a$) in all the metrics.

Analyzing the proposed method, we find that the correlation constraint ($h_p^C$) and the negative gradient constraint ($h_p^{NG}$) demonstrate comparable performance across various HI metrics. In contrast, the function-based constraint ($h_p^F$) exhibits relatively poorer performance in terms of mutual information and MAPE.

\begin{table}[h!] \small  
\begin{center}  
\begin{tabular}{ l  l  l  l  l  l  }
\hline \hline
\textbf{HI}	 \rule{0pt}{1.0\normalbaselineskip} & \textbf{Mon} &  \textbf{Tren} & \textbf{Prog} & \textbf{MutInf} & \textbf{MAPE} \\
\hline \hline
\multicolumn{6}{l}{Residual Method \rule{0pt}{1.0\normalbaselineskip}} \\
\hline 
$h_{r}^a$\rule{0pt}{1.0\normalbaselineskip} & 0.18\tiny(0.04) & 0.79\tiny(0.08) & 0.85\tiny(0.05) & 0.50\tiny(0.06) &24.4\tiny(4.3)  \\ [5pt]

$h_{r}^b$ & 0.33\tiny(0.03) & 0.91\tiny(0.03) & \textbf{0.98\tiny(0.01)} & 0.62\tiny(0.02) & 10.9\tiny(1.5)  \\ [5pt]

\hline 
\multicolumn{6}{l}{Proposed Method} \\
\hline 
\rowcolor{Gray}
$h_{p}^C$  \rule{0pt}{1.0\normalbaselineskip} & 0.33\tiny(0.02) & \textbf{0.98\tiny(0.00)} & 0.93\tiny(0.01) & \textbf{0.81\tiny(0.01)} &8.8\tiny(0.7) \\ [5pt]
\rowcolor{Gray}
$h_{p}^{NG}$   &\textbf{0.36\tiny(0.05)} & \textbf{0.98\tiny(0.00)} & 0.95\tiny(0.01) & \textbf{0.81\tiny(0.01)} & \textbf{8.5\tiny(1.6)}  \\ [5pt]
$h_{p}^{F}$ \rule{0pt}{1.0\normalbaselineskip}   &0.28\tiny(0.01) & 0.97\tiny(0.01) & 0.85\tiny(0.03) & 0.66\tiny(0.02) & 9.5\tiny(1.0)  \\ [5pt]
\hline 
\multicolumn{6}{l}{Supervised} \\
\hline 
$h_{s}$   \rule{0pt}{1.0\normalbaselineskip} & 0.40\tiny(0.03) &0.99\tiny(0.00)  & 0.94\tiny(0.01) & 0.79\tiny(0.01) & 8.2\tiny(1.4)\\[5pt]
\hline 
\multicolumn{6}{l}{Ground Truth} \\
\hline
$h_{gt}$ \rule{0pt}{1.0\normalbaselineskip} & 0.50 & 0.99 & 1.0 & 0.84&- 
\end{tabular}
\caption{Results of the HI criteria for the turbofan dataset. Mean and standard deviation values over 5 runs are provided.  $h_{r}^a$ - health index of AE residual method, $h_{r}^b$ - health index of regression residual method,  $h_{p}^C$ - health index of the proposed method using correlation constraint, $h_{p}^{NG}$ - health index of the proposed method using negative gradient constraint, $h_{p}^{F}$ - health index of the proposed method using functional constraint, $h_s$ - health index of supervised model, $h_{gt}$ - ground truth health index.}
 \label{table:ResultsHI_turbofan}
	\end{center}
\end{table}

It is worth noticing that the proposed methods consistently outperform the residual methods across all metrics, except for prognosability. This trend is further supported by the MAPE score, indicating that the proposed methods yield HIs more closely aligned with the ground truth compared to the residual methods. Furthermore, when considering a supervised model ($h_s$) for HI estimation as an alternative to our proposed unsupervised method, we observe only marginal performance enhancements.

\begin{figure*}[h!]%
    \centering
    \subfloat[\centering $h_r^a$]{{\includegraphics[scale = 0.8]{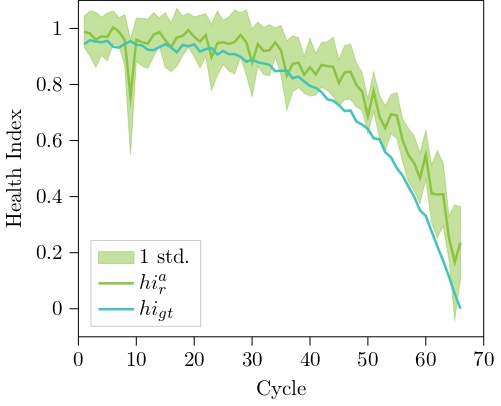} }}%
    \subfloat[\centering $h_r^b$]{{\includegraphics[scale = 0.8]{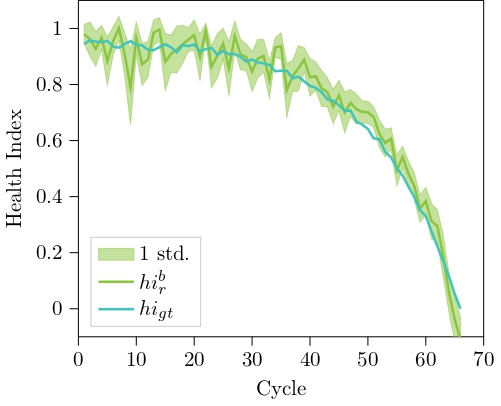} }}%

    \subfloat[\centering $h_{p}^C$]{{\includegraphics[scale = 0.8]{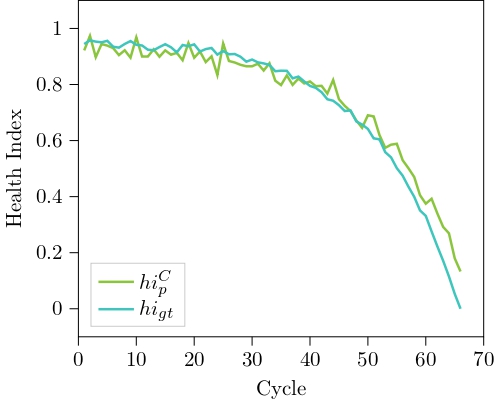} }}%
    \subfloat[\centering $h_{p}^{NG}$]{{\includegraphics[scale = 0.8]{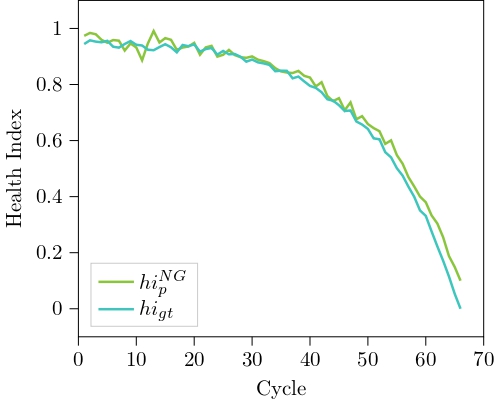} }}%

    \subfloat[\centering $h_{p}^F$]{{\includegraphics[scale = 0.8]{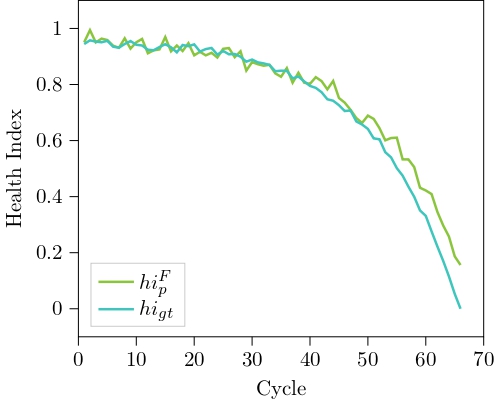} }}%
    \subfloat[\centering $h_{s}$]{{\includegraphics[scale = 0.8]{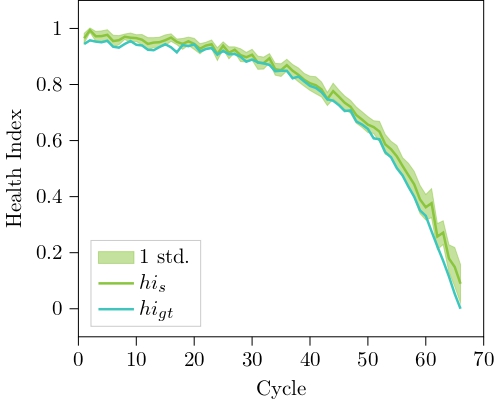} }}%
    \caption{Estimated HI of test unit 10 for the turbofan dataset. Residual methods (a) and (b). Proposed methods (c), (d), and (e). Supervised model (f).}%
    \label{fig:CMAPSS_HI_Visual}%
\end{figure*}

In addition to the quantitative evaluation of the HI metric, Figure \ref{fig:CMAPSS_HI_Visual} depicts the estimated HI for test unit 10 in the considered methods.  This unit is randomly selected for visualization purposes. The proposed method utilizing the negative gradient constraint (i.e., $h_{p}^{NG}$) results in an HI showing a closer match to the ground truth HI ($h_{gt})$ than the other methods considered. 

The prognostic prediction performance obtained when the prognostics model trained with the HI estimated with six evaluated methods is shown in Table \ref{table:ResultsRUL_turbofan}. The results show improved prognostics when integrating ground truth HI into the training data set. On average, the model shows a 32\% improvement compared to the baseline model, which underlines the importance of HI as a valuable source for the prediction of RUL.
Subsequently, the second most effective option is the supervised model, resulting in a 29\% performance enhancement. The proposed method yields comparable results, with the proposed negative gradient constraint showing a notable improvement of 28\%. It is worth noting that the inclusion of HI estimated by the residual method presents the least favorable scenario, yielding an improvement of 24\%.

\begin{table}[h!] \small  
	\begin{center}  
	\begin{tabular}{ lllll  }
		\hline \hline
		\textbf{Model} \rule{0pt}{1.0\normalbaselineskip}	& \textbf{MAE} &  \textbf{RMSE} & \textbf{MAPE} & \begin{tabular}[c]{@{}l@{}} \textbf{\% Average}\\ \textbf{Improvement}\end{tabular}   \\ 
		\hline \hline 

		\hline 
            \multicolumn{5}{l}{Baseline Model \rule{0pt}{1.0\normalbaselineskip}} \\
		\hline
            $G(X,W,t)$\rule{0pt}{1.0\normalbaselineskip} & 6.0\tiny(0.4) & 7.4\tiny(0.3) & 30.5\tiny(4.8) & -  \\ [5pt]
		\hline 
            \multicolumn{5}{l}{Residual Method } \\
		\hline
            $G(X,W,t,h_{re}^b)$\rule{0pt}{1.0\normalbaselineskip} & 4.9\tiny(0.2) & 6.7\tiny(0.1) & 16.1\tiny(1.2) & 24\%  \\ [5pt]
		\hline 
            \multicolumn{5}{l}{Proposed Method } \\
		\hline
            $G(X,W,t,h_{p}^C)$\rule{0pt}{1.0\normalbaselineskip} & \textbf{4.7\tiny(0.1)} & 6.9\tiny(0.2) & 15.0\tiny(0.4) & 26\% \\ [5pt]
            \rowcolor{Gray}
            $G(X,W,t,h_{p}^{NG})$ &4.9\tiny(0.1)  &\textbf{6.5\tiny(0.1)}  &\textbf{14.4\tiny(0.6)}&\textbf{28\%} \\ [5pt]

            $G(X,W,t,h_{p}^{F})$ \rule{0pt}{1.0\normalbaselineskip} &4.8\tiny(0.1)  &6.6\tiny(0.2)  &15.0\tiny(0.4)& 27\%\\ [5pt]
		\hline 
            \multicolumn{5}{l}{Supervised} \\
		\hline
            $G(X,W,t,h_{s})$\rule{0pt}{1.0\normalbaselineskip} &4.6\tiny(0.0)  &6.4\tiny(0.1) &15.6\tiny(0.7) &29\%  \\ [5pt]
		\hline 
            \multicolumn{5}{l}{Ground Truth} \\
		\hline
            $G(X,W,t,h_{gt})$\rule{0pt}{1.0\normalbaselineskip} &4.6\tiny(0.0)  & 6.3\tiny(0.1)  & 13.1\tiny(0.3) & 32\% 
	\end{tabular}
	\caption{Results of the prognostic prediction task for the turbofan dataset. Mean and standard deviation values over 5 runs are provided. $G$ - neural network, $X$ - sensor readings, $W$ -operating conditions, $t$ - cycles, $h_{r}^b$ - health index of regression residual method,  $h_{p}^C$ - health index of the proposed method using correlation constraint, $h_{p}^{NG}$ - health index of the proposed method using negative gradient constraint, $h_{p}^{F}$ - health index of the proposed method using functional constraint, $h_s$ - health index of supervised model, $h_{gt}$ - ground truth health index.}
	\label{table:ResultsRUL_turbofan}
	\end{center}
\end{table}

\subsubsection{In-Distribution Testing - Batteries}
\label{Results:Battery}

The HI metrics obtained from the six evaluated models are shown in Table \ref{table:ResultsHI_battery}. Comparing residual-based methods, both trained with CM data from the initial 100 random walk cycles of each training unit, we observe that the AE-based method outperforms the regression-based residual method in all the metrics.

Among the proposed methods, both the correlation constraint and the functional constraint exhibit superior performance, showcasing equivalent results. Conversely, the negative gradient constraint displays comparatively inferior performance across the HI criteria.

Compared to residual methods, the proposed methods incorporating correlation and functional constraints exhibit superior performance. Additionally, when comparing the proposed methods to the supervised model, an overall equivalency in performance is observed, albeit with notable distinctions. While the supervised model shows significantly improved results in terms of MAPE, the other criteria demonstrate a nearly identical performance.

\begin{table}[h!] \small  
	\begin{center}  
	\begin{tabular}{ llllll }
		\hline \hline
		\textbf{HI}	 \rule{0pt}{1.0\normalbaselineskip} & \textbf{Mon} &  \textbf{Tren}& \textbf{Prog} & \textbf{MutInf} & \textbf{MAPE} \\
		\hline \hline
            \multicolumn{6}{l}{Residual Method \rule{0pt}{1.0\normalbaselineskip}} \\
		\hline
            $h_{re}^a$ \rule{0pt}{1.0\normalbaselineskip}  & 0.66\tiny(0.07) &0.96\tiny(0.02) &0.78\tiny(0.01)  &0.53\tiny(0.04) &9.1\tiny(1.0) \\  [5pt]
            $h_{re}^b$  &0.50\tiny(0.07)  &0.64\tiny(0.02)  &0.66\tiny(0.01)  &0.34\tiny(0.01) & 11.0\tiny(0.2)  \\ [5pt]
		\hline 
            \multicolumn{6}{l}{Proposed Method} \\
		\hline
            \rowcolor{Gray}
            $h_{p}^{C}$  \rule{0pt}{1.0\normalbaselineskip}  & \textbf{0.86\tiny(0.04)}  &\textbf{0.99\tiny(0.00)}  & \textbf{0.83\tiny(0.02)} & \textbf{0.63\tiny(0.00)} & 7.3\tiny(0.6)\\ [5pt]
            $h_{p}^{NG}$  \rule{0pt}{1.0\normalbaselineskip}  &0.39\tiny(0.08)  &0.84\tiny(0.09)  &0.83\tiny(0.11)  &0.43\tiny(0.07)&9.6\tiny(1.0)  \\ [5pt]
            \rowcolor{Gray}
            $h_{p}^{F}$  \rule{0pt}{1.0\normalbaselineskip}  &0.73\tiny(0.04)  &0.98\tiny(0.00)  &0.82\tiny(0.01)  &0.62\tiny(0.00)&\textbf{7.0\tiny(0.5)}  \\ [5pt]

		\hline 
            \multicolumn{6}{l}{Supervised} \\
		\hline
            $h_{s}$   \rule{0pt}{1.0\normalbaselineskip} & 0.94\tiny(0.08) &0.98\tiny(0.08) &0.88\tiny(0.08)  & 0.62\tiny(0.02)&3.1\tiny(0.2)\\[5pt]

		\hline 
            \multicolumn{6}{l}{Ground Truth} \\
		\hline
            $h_{gt}$ \rule{0pt}{1.0\normalbaselineskip} & 0.97 & 1.00 & 0.97&0.65&- 
	\end{tabular}
 	\caption{Results of the HI criteria for the battery dataset.}
        \label{table:ResultsHI_battery}
	\end{center}
\end{table}

\begin{figure*}[h!]%
    \centering
    \subfloat[\centering $h_r^a$]{{\includegraphics[scale = 0.8]{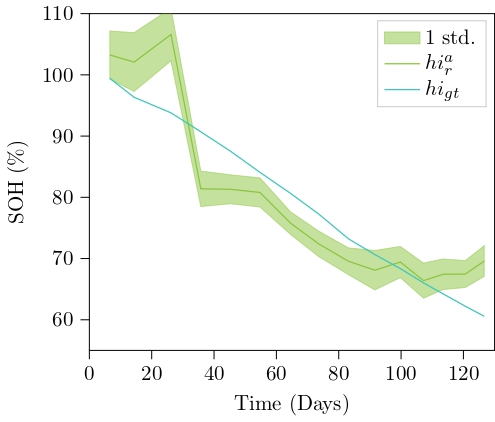} }}%
    \subfloat[\centering $h_r^b$]{{\includegraphics[scale = 0.8]{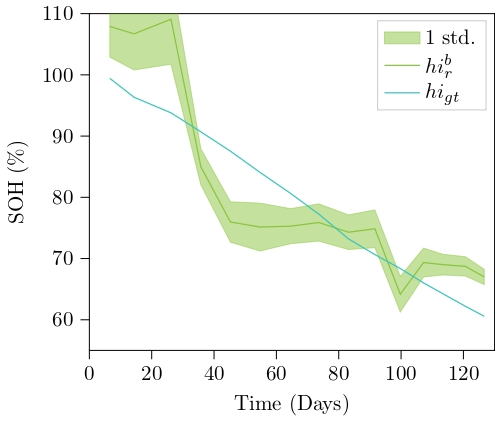} }}%

    \subfloat[\centering $h_{p}^C$]{{\includegraphics[scale = 0.8]{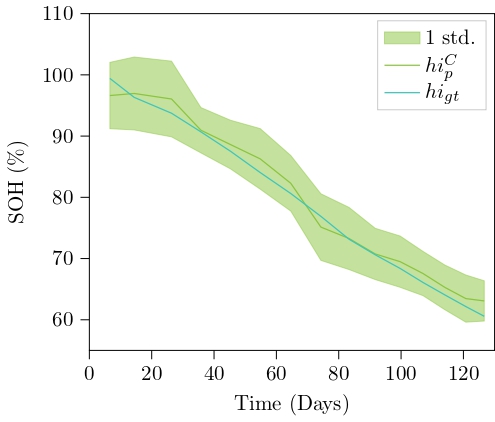} }}%
    \subfloat[\centering $h_{p}^{NG}$]{{\includegraphics[scale = 0.8]{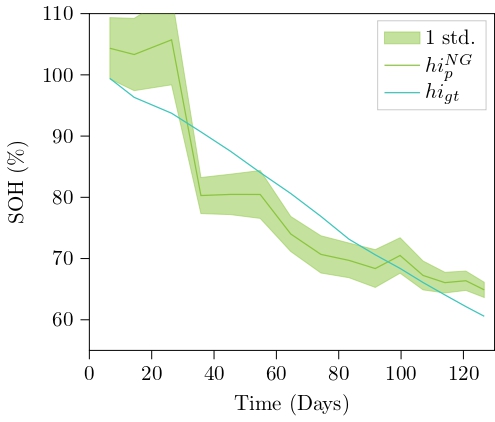} }}%

    \subfloat[\centering $h_{p}^F$]{{\includegraphics[scale = 0.8]{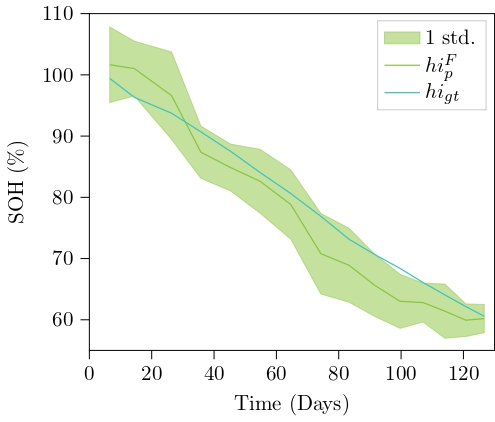}}}%
    \subfloat[\centering $h_{s}$]{{\includegraphics[scale = 0.8]{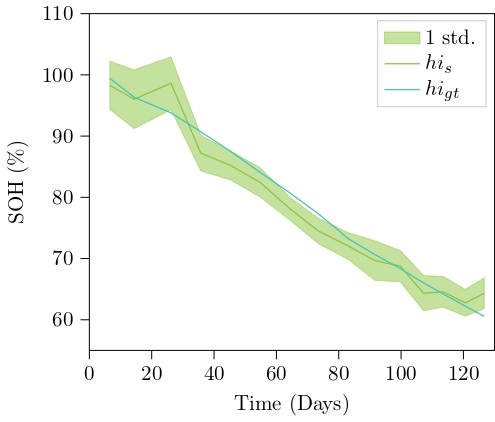} }}%
    \caption{Estimated HI of test unit 20 for the battery dataset. Residual methods (a) and (b). Proposed methods (c), (d), and (e). Supervised model (f).}%
    \label{fig:Battery_HI_Visual}%
\end{figure*}

Figure \ref{fig:Battery_HI_Visual} depicts the estimated HI for test battery 20 in the considered methods. This battery is randomly selected for visualization purposes. The HI estimated by the supervised model seems to be a closer match to the ground truth HI than HIs estimated by the other methods. The proposed methods employing the correlation or functional constraint are the next best choice. 

The prognostic prediction performance obtained when the prognostics model trained with the HI estimated with six evaluated methods is shown in Table \ref{table:ResultsRUL_battery}. The results show improved prognostics when integrating ground truth HI into the training data set. On average, the model shows a 54\% improvement compared to the baseline model. Subsequently, the second most effective option is the supervised model, resulting in a 26\% performance enhancement. The results of the proposed method yield comparable results, with the proposed functional constraint showing a notable improvement of 25\%. The other two constraints, namely correlation and negative gradient, lead to 20\% and 11\%, respectively. Using the residual method leads to 14\% improvement.

\begin{table}[!h] \small  
	\begin{center}  
	\begin{tabular}{ lllll  }
		\hline \hline
		\textbf{Model} \rule{0pt}{1.0\normalbaselineskip}	& \textbf{MAE} &  \textbf{RMSE} & \textbf{MAPE} & \begin{tabular}[c]{@{}l@{}} \textbf{\% Average}\\ \textbf{Improvement}\end{tabular}    \\ 
		\hline \hline 
            \multicolumn{5}{l}{Baseline \rule{0pt}{1.0\normalbaselineskip}} \\
		\hline
            $G(X,W,t)$\rule{0pt}{1.0\normalbaselineskip} &165\tiny(15) &206\tiny(17) &256\tiny(16) & -  \\ [5pt]
		\hline 
            \multicolumn{5}{l}{Residual Method} \\
		\hline
            $G(X,W,t,h_{re}^a)$\rule{0pt}{1.0\normalbaselineskip} &144\tiny(11) &180\tiny(14) &215\tiny(19) & 14\%   \\ [5pt]
		\hline 
            \multicolumn{5}{l}{Proposed Method} \\
		\hline         
            $G(X,W,t,h_{p}^C)$\rule{0pt}{1.0\normalbaselineskip} &134\tiny(5) &166\tiny(5) &202\tiny(13) & 20\%   \\ [5pt]

            $G(X,W,t,h_{p}^{NG})$ &150\tiny(5) &185\tiny(5) &222\tiny(16)& 11\%     \\ [5pt]
            \rowcolor{Gray}
            $G(X,W,t,h_{p}^{F})$ &\textbf{126\tiny(10)} &\textbf{161\tiny(13)} &\textbf{180\tiny(14)}& \textbf{25\%}     \\ [5pt]
		\hline 
            \multicolumn{5}{l}{Supervised } \\
		\hline
            $G(X,W,t,h_{s})$\rule{0pt}{1.0\normalbaselineskip} &134\tiny(5) &167\tiny(8) & 153\tiny(15)& 26\%   \\ [5pt]
		\hline 
            \multicolumn{5}{l}{Ground Truth} \\
		\hline
            $G(X,W,t,h_{gt})$\rule{0pt}{1.0\normalbaselineskip}&96\tiny(8) &113\tiny(9) &67\tiny(7) & 54\%    
	\end{tabular}
	\caption{Results of the prognostic prediction task for the battery dataset.}
	\label{table:ResultsRUL_battery}
	\end{center}
\end{table}

\subsection{Out-of-Distribution Testing }
\label{Results:Abalation}

The accuracy and reliability of prognostic prediction techniques hinge on the quality and representativeness of available time-to-failure data. As a result, these techniques may exhibit reduced performance when applied to data from new units operating under conditions distinct from those in the training set \cite{nejjar2023domain}, leading to an Out-of-Distribution (OOD) scenario. To assess the capabilities of the proposed method under such OOD scenarios, we intentionally designed challenging scenarios for each dataset.

For the turbofan dataset, we create an OOD scenario by considering different flight classes. In particular, we train models with short-flight class data and conduct tests on medium to long-flight classes. The specific training and testing units are detailed in Table \ref{table:OODTurbofan}.

\begin{table}[h!]
\centering
\begin{tabular}{l|l|l}
 \hline \hline
Flight Class \rule{0pt}{1.0\normalbaselineskip}   & Training Units & Testing Units \\
 \hline \hline 
Short      \rule{0pt}{1.0\normalbaselineskip} & U1, U5, U9, U12, U14  & -  \\
Medium   & - &  U2, U3, U4, U7, U15           \\
Long  & - & U6, U8, U10, U11, U13  \\
\end{tabular}
\caption{Out-of-distribution training set-up for N-CMAPSS dataset.}
\label{table:OODTurbofan}
\end{table}

For the battery dataset, we create an OOD scenario by considering different load profiles. We train models using uniform load data and conduct tests using low-skew and high-skew data. For more information on the training/testing units see Table \ref{table:OODBattery}.

\begin{table}[h!]
\centering
\begin{tabular}{l|l|l}
 \hline \hline
Load profile \rule{0pt}{1.0\normalbaselineskip}   & Training Batteries & Testing Batteries \\
 \hline \hline 
Uniform     \rule{0pt}{1.0\normalbaselineskip}       & RW1,  RW4 - RW8 & -  \\
Skewed High  & - & RW17, RW19, RW20 \\
Skewed Low  & - & RW13 - RW16 \\
\end{tabular}
\caption{Out-of-distribution training set-up for NASA battery dataset.}
\label{table:OODBattery}
\end{table}

\subsubsection{OOD Testing - Turbofan}
\label{Results:Turbofan_OOD}
Table \ref{table:ResultsHI_turbofan_ood} illustrates the HI metrics in the out-of-distribution scenario. The proposed method, incorporating the functional constraint, surpasses the residual-based method across all metrics. However, the use of the correlation constraint does not yield improved performance compared to the residual method.

Moreover, when evaluating a supervised model for HI estimation, only marginal performance enhancements are observed in comparison to both the residual method and the proposed method with the correlation constraint. The supervised model demonstrates inferior performance across all metrics when compared to the proposed method utilizing the functional constraint.

\begin{table}[h!] \small  
	\begin{center}  
	\begin{tabular}{ l  l  l  l  l  l  }
		\hline \hline
		\textbf{HI}	 \rule{0pt}{1.0\normalbaselineskip} & \textbf{Mon} &  \textbf{Tren} & \textbf{Prog} & \textbf{MutInf} & \textbf{MAPE} \\
		\hline \hline
            \multicolumn{6}{l}{Residual Method \rule{0pt}{1.0\normalbaselineskip}} \\
		\hline 
             $h_{r}^b$  \rule{0pt}{1.0\normalbaselineskip} & 0.12\tiny(0.03) &0.68\tiny(0.05) & 0.86\tiny(0.03) & 0.45\tiny(0.07) & 25.1\tiny(3.6) \\ [5pt]
		\hline 
            \multicolumn{6}{l}{Proposed Method} \\
		\hline 
            $h_{p}^C$  \rule{0pt}{1.0\normalbaselineskip} & 0.10\tiny(0.03) & 0.75\tiny(0.12) & 0.74\tiny(0.13) & 0.57\tiny(0.10) &34.2\tiny(5.9) \\ [5pt]
            \rowcolor{Gray}
            $h_{p}^F$  \rule{0pt}{1.0\normalbaselineskip} & \textbf{0.16\tiny(0.03)} & \textbf{0.91\tiny(0.06)} & \textbf{0.89\tiny(0.05)} & \textbf{0.68\tiny(0.06)} &\textbf{16.6\tiny(3.5)} \\ [5pt]
		\hline 
            \multicolumn{6}{l}{Supervised } \\
		\hline 
            $h_{s}$   \rule{0pt}{1.0\normalbaselineskip} & 0.11\tiny(0.04) &0.80\tiny(0.05)  & 0.88\tiny(0.04) & 0.55\tiny(0.06) & 22.8\tiny(4.4)\\[5pt]
		\hline 
            \multicolumn{6}{l}{Ground Truth} \\
		\hline
            $h_{gt}$ \rule{0pt}{1.0\normalbaselineskip} & 0.53 & 0.99 & 1.0 & 0.84&- 
	\end{tabular}
	\caption{Results of the HI criteria for the turbofan dataset under out-of-distribution scenario.}
\label{table:ResultsHI_turbofan_ood}
	\end{center}
\end{table}

For the prognostics prediction task in the out-of-distribution scenario, we decided to cap the maximum RUL value to 60 cycles to prevent large prediction errors for healthy units. The results are given in Table \ref{table:ResultsRUL_turbofan_ood}. 

The results show the improvement in prognostics when integrating ground truth HI into the training data set. On average the model shows  47\% improvement when incorporating the ground truth HI compared to the baseline model. The second most effective option was integrating the HI estimated by the supervised model. However, the inclusion of HI estimated by the proposed method, utilizing a functional constraint, showed a similar performance boost (41\% compared to 37\%). Comparatively, the inclusion of HI estimated by the residual method and the proposed method employing the correlation constraint result in marginal improvements of 5\% and 9\%, respectively.

\begin{table}[!h] \small  
	\begin{center}  
	\begin{tabular}{ lllll  }
		\hline \hline
		\textbf{Model} \rule{0pt}{1.0\normalbaselineskip}	& \textbf{MAE} &  \textbf{RMSE} & \textbf{MAPE} & \begin{tabular}[c]{@{}l@{}} \textbf{\% Average}\\ \textbf{Improvement}\end{tabular}   \\ 
		\hline \hline

            \multicolumn{5}{l}{Baseline Model \rule{0pt}{1.0\normalbaselineskip}} \\
		\hline
            $G(X,W,t)$\rule{0pt}{1.0\normalbaselineskip} & 11.2\tiny(1.4) & 13.9\tiny(1.4) & 56.1\tiny(14.4) & -  \\ [5pt]
		\hline 
            \multicolumn{5}{l}{Residual Method } \\
		\hline

            $G(X,W,t,h_{r}^b)$\rule{0pt}{1.0\normalbaselineskip} & 11.2\tiny(1.1) & 14.4\tiny(1.2) & 44.9\tiny(10.3) & 5\%  \\ [5pt]
            
		\hline 
            \multicolumn{5}{l}{Proposed Method } \\
		\hline
            $G(X,W,t,h_{p}^C)$\rule{0pt}{1.0\normalbaselineskip} &10.4\tiny(0.9)  &12.8\tiny(1.1) &50.0\tiny(12.7) &9\%  \\ [5pt]
            \rowcolor{Gray}
            $G(X,W,t,h_{p}^F)$\rule{0pt}{1.0\normalbaselineskip} &\textbf{6.8\tiny(0.6)}  &\textbf{9.1\tiny(1.0)}&\textbf{35.8\tiny(7.1)} &\textbf{37\%}  \\ [5pt]

		\hline 
            \multicolumn{5}{l}{Supervised } \\
		\hline
            $G(X,W,t,h_{s})$\rule{0pt}{1.0\normalbaselineskip} &7.0\tiny(1.0)  &9.1\tiny(1.4) &27.4\tiny(5.0) &41\%  \\ [5pt]

		\hline 
            \multicolumn{5}{l}{Ground Truth } \\
		\hline
            $G(X,W,t,h_{gt})$\rule{0pt}{1.0\normalbaselineskip} &6.2\tiny(0.6)  & 8.5\tiny(0.6)  & 23.3\tiny(3.4) & 47\% 
	\end{tabular}
	\caption{Results of the prognostics prediction task for the turbofan dataset under out-of-distribution scenario.}
	\label{table:ResultsRUL_turbofan_ood}
	\end{center}
\end{table}

\subsubsection{OOD Testing - Batteries} 
\label{Results:Battery_OOD}

Table \ref{table:ResultsHI_Battery_ood} presents the HI metrics for the out-of-distribution scenario concerning the battery dataset. In comparison to the residual method, the proposed method using the functional constraint exhibits superior performance in trendability, mutual information, and MAPE. However, the proposed method using the correlation constraint performs worse than the residual method across all metrics except MAPE.

The HI metrics show a minimal performance gap between the supervised model and the proposed method utilizing the functional constraint. The proposed method with the functional constraint outperforms the supervised model in terms of trendability and mutual information, but scores lower in terms of monotonicity and MAPE. 

\begin{table}[h!] \small  
	\begin{center}  
	\begin{tabular}{ llllll  }
		\hline \hline
		\textbf{HI}	 \rule{0pt}{1.0\normalbaselineskip} & \textbf{Mon} &  \textbf{Tren} & \textbf{Prog} & \textbf{MutInf} & \textbf{MAPE} \\
		\hline 
            \hline
            \multicolumn{6}{l}{Residual Method \rule{0pt}{1.0\normalbaselineskip}} \\
		\hline 
            $h_{r}^a$   \rule{0pt}{1.0\normalbaselineskip} & \textbf{0.76\tiny(0.06)} &0.97\tiny(0.00)  &\textbf{0.85\tiny(0.04)}  & 0.58\tiny(0.05) & 9.5\tiny(0.5)\\  [5pt]
            \hline
            \multicolumn{6}{l}{Proposed Method \rule{0pt}{1.0\normalbaselineskip}} \\
		\hline 
            $h_{p}^C$   \rule{0pt}{1.0\normalbaselineskip} & 0.41\tiny(0.11) & 0.87\tiny(0.10) & 0.71\tiny(0.04) & 0.45\tiny(0.13) &9.0\tiny(0.9)\\  [5pt]
            \rowcolor{Gray}
            $h_{p}^{F}$  \rule{0pt}{1.0\normalbaselineskip} &0.61\tiny(0.07) & \textbf{0.97\tiny(0.01)} & 0.79\tiny(0.02) & \textbf{0.61\tiny(0.02)} & \textbf{8.0\tiny(0.3)}  \\ [5pt]
            
            \hline
            \multicolumn{6}{l}{Supervised \rule{0pt}{1.0\normalbaselineskip}} \\
		\hline 
            $h_{s}$   \rule{0pt}{1.0\normalbaselineskip} &0.71\tiny(0.04) &0.93\tiny(0.04)  & 0.77\tiny(0.13) & 0.58\tiny(0.04) & 6.4\tiny(1.3)\\[5pt]
            \hline
            \multicolumn{6}{l}{Ground Truth \rule{0pt}{1.0\normalbaselineskip}} \\
		\hline 
            $h_{gt}$ \rule{0pt}{1.0\normalbaselineskip} & 0.98 & 0.99 & 0.97 & 0.68&- 
	\end{tabular}
 	\caption{Results of the HI criteria for the battery dataset under out-of-distribution scenario.}
\label{table:ResultsHI_Battery_ood}
	\end{center}
\end{table}

For the prognostics prediction task in the out-of-distribution scenario, we decided to cap the maximum RUL value to 300 cycles to prevent large prediction errors. The results are given in Table \ref{table:ResultsRUL_battery_ood}. 

Introducing the ground truth HI significantly enhances RUL prediction, achieving an average improvement of 66\%. The next best approach involves integrating an HI estimated through the proposed method using the function constraint, resulting in a notable performance increase of 31\%. Incorporating an HI estimated by the supervised model closely follows, showcasing an improvement of 28\%. Conversely, integrating an HI estimated by the proposed method using the correlation constraint or the residual method yields more modest improvements, increasing performance by 23\% and 19\%, respectively.

\begin{table}[!h] \small  
	\begin{center}  
	\begin{tabular}{ lllll  }
		\hline \hline
		\textbf{Model} \rule{0pt}{1.0\normalbaselineskip}	& \textbf{MAE} &  \textbf{RMSE} & \textbf{MAPE} & \begin{tabular}[c]{@{}l@{}} \textbf{\% Average}\\ \textbf{Improvement}\end{tabular}   \\ 
		\hline  \hline 
            \multicolumn{5}{l}{Baseline Model \rule{0pt}{1.0\normalbaselineskip}} \\
		\hline
            $G(X,W,t)$\rule{0pt}{1.0\normalbaselineskip} &52\tiny(8) &85\tiny(18) &142\tiny(37) & -  \\ [5pt]
		\hline 
            \multicolumn{5}{l}{Residual Method } \\
		\hline
            $G(X,W,t,h_{r}^a)$\rule{0pt}{1.0\normalbaselineskip} & 52\tiny(5)&68\tiny(6) &89\tiny(12) & 19\%   \\ [5pt]  
		\hline 
            \multicolumn{5}{l}{Proposed Method } \\
		\hline
            $G(X,W,t,h_{p}^C)$\rule{0pt}{1.0\normalbaselineskip} &46\tiny(3) &71\tiny(3) &83\tiny(6) & 23\%   \\ [5pt]
            \rowcolor{Gray}
            $G(X,W,t,h_{p}^{F})$ \rule{0pt}{1.0\normalbaselineskip} &\textbf{41\tiny(6)} &\textbf{64\tiny(9)} &\textbf{77\tiny(6)}& \textbf{31\%}     \\ [5pt]
		\hline 
            \multicolumn{5}{l}{Supervised } \\
		\hline
            $G(X,W,t,h_{s})$\rule{0pt}{1.0\normalbaselineskip} &43\tiny(5) &69\tiny(5) &74\tiny(4) & 28\%   \\ [5pt]
		\hline 
            \multicolumn{5}{l}{Ground Truth } \\
		\hline
            $G(X,W,t,h_{gt})$\rule{0pt}{1.0\normalbaselineskip}&21\tiny(1) &36\tiny(4) &26\tiny(7) & 66\%   
	\end{tabular}
	\caption{Results of the prognostics prediction task for the battery dataset under out-of-distribution scenario.}
	\label{table:ResultsRUL_battery_ood}
	\end{center}
\end{table}

\subsection{Ablation Study: Impact of Hybridization Techniques}
\label{sec:Ablation}

This section expands on the prior analysis by conducting an ablation study to show the advantages of incorporating multiple hybridization strategies in the proposed method. Specifically, the proposed method incorporates two hybridization strategies: inductive bias and learning bias. To assess the independent impact of each hybridization strategy on accurate HI estimation, we systematically eliminate each bias from the model.

Initially, we examine the effect of removing the learning bias by setting the parameter $\lambda$, which controls the importance of the additional constraint, to 0. Subsequently, we investigate the effect of eliminating the inductive bias while preserving the learning bias by employing a convolutional AE where W and X serve as input to reconstruct X. The architecture mirrors that of the proposed method, with the distinction that the operating conditions are input to the encoder and the decoder  (i.e., see UL method in Figure \ref{fig:Methods_overview}). Lastly, we eliminate both learning bias and inductive bias, essentially creating a fully data-driven unsupervised model. This is achieved by employing the same architecture as in the second case but omitting the additional constraint term from the objective function. 

\begin{figure*}[h!]%
    \centering

    \subfloat[\centering Proposed method]{{\includegraphics[scale = 0.8]{HI_C_turbofan.jpg} }}%
    \subfloat[\centering With inductive bias, without learning bias]{{\includegraphics[scale = 0.8]{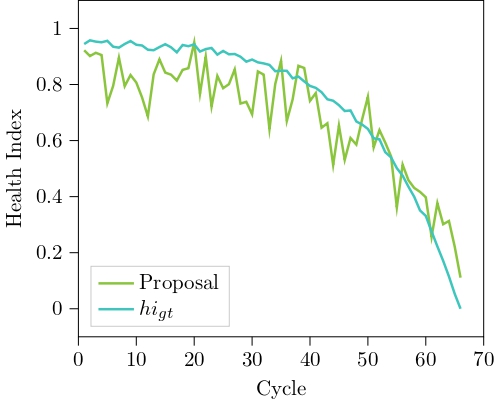} }}%

    \subfloat[\centering Without inductive bias, with learning bias]{{\includegraphics[scale = 0.8]{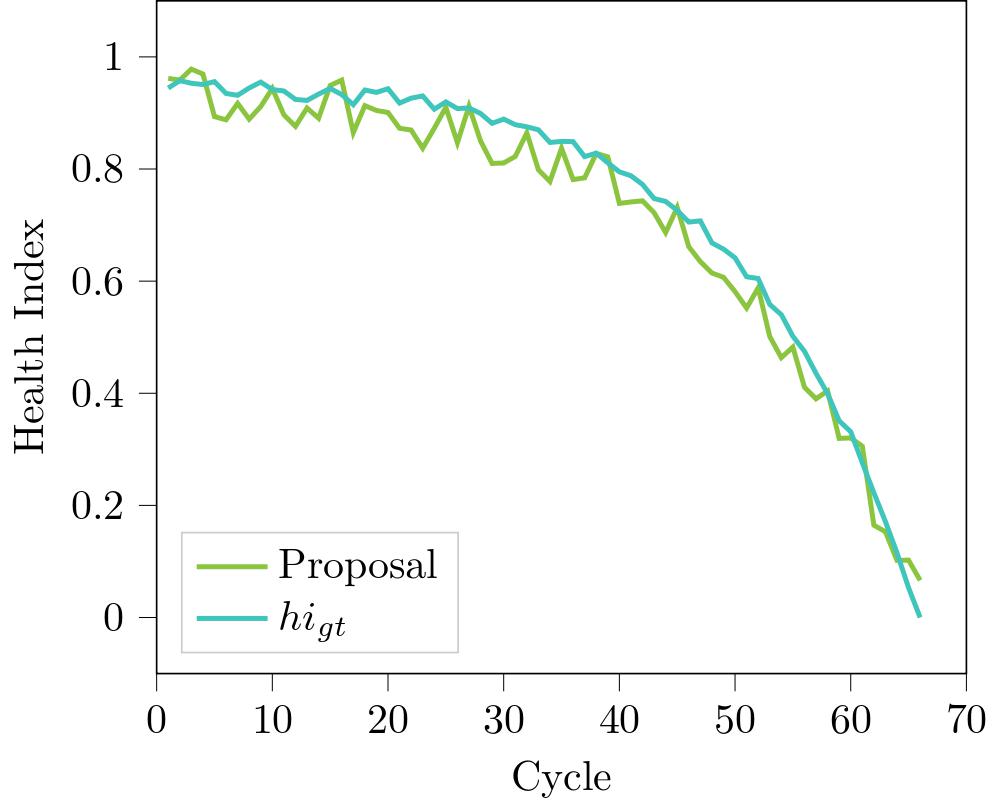} }}%
    \subfloat[\centering Without inductive bias, without learning bias]{{\includegraphics[scale = 0.8]{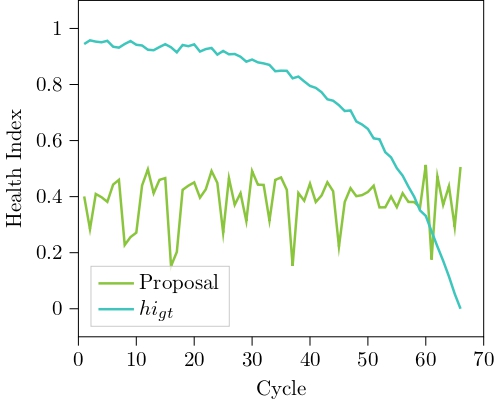} }}%

    \caption{Estimated HI of test unit 10 for the turbofan dataset. Proposed method (a) Proposed method. (b) Proposed method without learning bias. (c) Proposed method without inductive bias. (d) Data-driven model.}%
    \label{fig:HI_Abalation}%
\end{figure*}

The experiments focus on the in-distribution case of the turbofan dataset; the resulting HI metrics are shown in Table \ref{table:ResultsAbalation}. In the initial scenario of eliminating the learning bias, a substantial decline in HI metrics is evident compared to the proposed method. In the subsequent case of removing the inductive bias, we present the outcomes of incorporating the correlation constraint. The results indicate a less pronounced decrease in HI metrics compared to the proposed method, with the most significant impact observed in Mutual Information and MAPE. Finally, we demonstrate that integrating no prior knowledge into the model results in the worst estimation of the HI. Figure \ref{fig:HI_Abalation} visually represents the estimated HIs for each ablation experiment.

\begin{table}[h!] \small  
\begin{center}  
\begin{tabular}{ lllll  }
\hline \hline
\textbf{Mon} \rule{0pt}{1.0\normalbaselineskip} &  \textbf{Tren} & \textbf{Prog} & \textbf{MutInf} & \textbf{MAPE} \\
\hline  \hline
\multicolumn{5}{c}{Proposed Method \rule{0pt}{1.0\normalbaselineskip}} \\
\hline 
0.36\tiny(0.05) \rule{0pt}{1.0\normalbaselineskip} & 0.98\tiny(0.00) & 0.95\tiny(0.01) & 0.81\tiny(0.01) & 8.5\tiny(1.6) \\ [5pt] \hline
\multicolumn{5}{c}{With Inductive Bias and without Learning Bias} \\
\hline 
0.11\tiny(0.02) \rule{0pt}{1.0\normalbaselineskip} & 0.77\tiny(0.06) & 0.93\tiny(0.03) & 0.41\tiny(0.08) & 26.4\tiny(6.5) \\ [5pt] \hline
\multicolumn{5}{c}{Without Inductive Bias and with Learning Bias} \\
\hline 
0.31\tiny(0.03) \rule{0pt}{1.0\normalbaselineskip} & 0.98\tiny(0.00) & 0.95\tiny(0.02) & 0.69\tiny(0.01) & 10.7\tiny(2.4) \\  \hline
\multicolumn{5}{c}{Without Inductive Bias and Learning Bias} \\
\hline 
0.05\tiny(0.01) \rule{0pt}{1.0\normalbaselineskip} & 0.01\tiny(0.01) & 0.03\tiny(0.03) & 0.03\tiny(0.01)& 89.6\tiny(6.0) 
\end{tabular}
\caption{Results of the HI criteria for the turbofan dataset ablation study.}
\label{table:ResultsAbalation}
\end{center}
\end{table}

\section{Discussion}
\label{Discussion}
This research aimed at achieving a reliable hybrid method for estimating the HI of diverse complex systems. In pursuit of this objective, we proposed a hybrid unsupervised method for HI estimation based on general knowledge. To compensate for the low informative nature of the prior knowledge, we opted to combine multiple hybridization strategies. 

To validate the generality of our proposed method, we conducted evaluations on two distinct case studies, each characterized by different degradation dynamics and their respective manifestations in observable sensor readings. The main findings are summarized as follows:

\begin{itemize}
    \item The proposed method outperformed the industry standard residual method in both case studies. Furthermore, the performance of the proposed method was on par with that of a supervised model. This suggests that incorporating general knowledge contributes significantly to the method's performance across diverse systems.

    \item Among the tested soft constraints, the functional constraint emerged as the most effective choice for both case studies. The functional constraint uses the most system-specific knowledge and compels the latent space of the model to adhere to specific values. 

    \item The advantage of the correlation and negative gradient constraints varied depending on the case study. In the case of turbofan engines characterized by highly non-linear degradation, the negative gradient constraint proved more effective. Conversely, for batteries exhibiting linear degradation and health recovery aspects, the correlation constraint demonstrated better performance.

    \item Ablation analysis (Section \ref{sec:Ablation}) shows that both hybridization strategies are effective for HI estimation, but the combination of the two is advantageous.
    \end{itemize}

The aforementioned findings show that our proposed method has good generalization since it can accurately estimate the HI of various systems. However, it is important to acknowledge that the proposed method does have certain limitations. We outline these limitations below.

\begin{itemize}
    \item The proposed method is tailored for systems characterized by failure modes predominantly driven by cycle loading. In instances where various factors drive the system's failure mechanism, the adaptability of the soft constraint employed in our method may require further consideration.
    
    
    \item The proposed method was only evaluated in case studies with continuous degradation. Exploring the suitability of our proposed method for systems exhibiting abrupt failures is a subject of future investigation.

    \item The evaluation of the proposed method involved case studies with run-to-failure data. Although our method applies to censored data, the interpreted meaning of the estimated HI differs. In instances where no failures are observed (HI = 0), the estimated HI is normalized with reference to the most degraded unit observed.
\end{itemize}

\section{Conclusion}
\label{Conclusion}

This work proposes an unsupervised hybrid method for HI estimation leveraging general knowledge and two hybridization strategies. The proposed method features two design features: 1) a novel network architecture of a convolutional AE preserving the causal relationships among sensor readings, operating conditions, and degradation within complex systems, and 2) the incorporation of soft constraints within the loss function derived from general knowledge of the degradation process, guiding the AE to infer degradation in its latent space. 

In an extensive analysis involving turbofan engines and batteries, both in-distribution and out-of-distribution testing scenarios, we demonstrated that this hybrid method, grounded in generalized knowledge, has wide applicability across diverse systems.

The effectiveness and generalization capabilities of the proposed method were demonstrated in comparative analysis involving alternative HI estimation methods. The evaluation encompassed both HI metrics and the utility of the HI for RUL prognostics. The proposed method consistently outperformed the industry standard residual method in all experimental setups. Notably, the performance gap between our approach and fully supervised models was minimal, particularly in RUL prediction tasks. For instance, in the turbofan dataset, both our method and the supervised model improved RUL predictions by approximately 28\%. Similarly, in the battery dataset, both methods yielded approximately 25\% improvement. The results emphasize the importance of integrating knowledge into neural networks, showcasing the informative potential embedded in such knowledge.

For future research, we plan to expand the application of our method beyond turbofan engines and batteries to include other critical systems, such as bearings. Furthermore, a key focus in future research will be the exploration of optimal strategies for effectively leveraging HI for RUL prognostics.

\section{Declaration of generative AI and AI-assisted technologies in the writing process}
During the preparation of this work, the author(s) used ChatGPT (3.5) in order to improve the readability and language of the text. After using this tool/service, the author(s) reviewed and edited the content as needed and take(s) full responsibility for the content of the publication.

\appendix
\section{Additive Causal Model}
\label{sec:ANM}

\edit{This section is focused on the domain of causal inference and presents a common methodology for uncovering causal relationships from observational data. We demonstrate the results of applying such methodology for the turbofan case study.}

\edit{
To facilitate clarity, we begin by introducing causal notation. Initially, our attention is directed towards scenarios characterized by causal models involving only two variables. We then follow with a brief overview of the procedure for a multivariate case. In the context of the bivariate scenario a structural causal model (SCM) consists of two assignments: }
\edit{
\begin{align}
        C& :=f_1(\epsilon_1) \\
        E& := f_2(C,\epsilon_2)
\end{align}
}
\edit{
where $\epsilon_1, \epsilon_2$ are jointly independent noise variables and $f_1,f_2$ are deterministic functions. In this model, we denote the random variables $C$ as the cause and $E$ as the effect. Furthermore, we refer to the causal graph $C\rightarrow E$ if $C$ is a direct cause of $E$. }

\edit{
Determining the causal direction, even in a bivariate scenario, is challenging \cite{mooij2016distinguishing}.  A recognized method for establishing causal direction is the non-linear additive noise model (ANM), as introduced in \cite{hoyer2008nonlinear}. In general, if $C$ is a direct cause of $E$, then it is intuitive to model the relationship as:}
\edit{
\begin{equation}
    E = f(C) +\epsilon \quad C \perp \epsilon
\end{equation}
}
\edit{
where $f(\cdot)$ is an arbitrary nonlinear function and $\epsilon$ is the independent noise variable. The assertion that $C$ is independent of $\epsilon$ ($C \perp \epsilon$) relies on several assumptions, including the absence of hidden common causes between  $C$ and $E$ and no feedback loops between the two (i.e., no $C \leftrightarrow E$ interaction).}

\edit{
The non-linear ANM can be effectively applied to observational data in practical settings. Given two random variables $C$ and $E$, the approach involves estimating the conditional expectation $\mathbb{E}(E|C)$ through regression analysis, followed by testing the independence of the residuals $E - \mathbb{E}(E|C)$ and $C$. Since the causal direction is typically unknown beforehand, it is necessary to test both possible causal directions. }

\edit{
The extension to the multivariate case is discussed in detail in \cite{hoyer2008nonlinear}. Briefly,  for each potential causal structure, represented by a directed acyclic graph (DAG) $G_i$, the procedure involves conducting a nonlinear regression for each variable against its parent variables. Subsequently, we test whether the resulting residuals are mutually independent. If any independence test is rejected, $G_i$ is rejected. However, depending on the significance levels for rejecting and accepting independence, the ANM may indicate causality in both directions, no direction, or only one direction. }

\edit{
To address cases where these tests are inconclusive, \cite{peters2017elements} propose an alternative method based on the variance score of residuals. This method evaluates causality by assigning a higher score to models where the variance of the residuals is smaller, indicating a better fit. Such scores help in making definitive decisions about the causal direction. The procedure is outlined in \cite{peters2017elements} and summarized in Algorithm \ref{ANM}.}

\begin{figure}[h!]
  \centering
\begin{minipage}{0.9\textwidth}
\begin{algorithm}[H]
\caption{\edit{General procedure to find the optimal causal structure graph $G_{opt}$.} }
\label{ANM}
\hspace*{\algorithmicindent} \textbf{Input:}
Observational data $\mathcal{D}_N=\{V_j\}_j^N$, variables $V_j$
\begin{algorithmic}[1]
\STATE Construct all possible DAG $G$ with $d$ structural assignments $V_j = f_j(PA(V_j),\epsilon_j) \quad j = 1,...,d$, where $PA(V_j)$ are the parents of $V_j$
\STATE For each graph structure $G_k$ regresses each variable $V_j$ on its parents $PA(V_j)$
\STATE Obtain residuals $R_j = V_j - \hat{f}_{j}(PA(V_j))$ 
\STATE Obtain a score  $\log p(\mathcal{D}|G_k) = \sum_{j=1}^{d}-\log(\text{var}(R_j))$
\STATE Obtain most probable causal graph $G_{opt} = \arg \max_{k} \{\log p(\mathcal{D}|G_k)\} $
\end{algorithmic}
\end{algorithm}
\end{minipage}
\end{figure}

\edit{
For the turbofan case study, we utilize the methodology given in Algorithm \ref{ANM} specifically designed for the multivariate scenario. However, due to the large number of variables involved, including 14 sensor readings denoted as X, 4 operating conditions denoted as W, and the system health indicator Z, the exploration of all possible DAGs becomes impractically large.}

\edit{
To address this challenge, we introduce simplifying assumptions. Firstly, we assume mutual independence among the sensor readings $X$ ($X_i \perp X_j, \forall i\neq j$) and mutual independence among the operating conditions $W$. This allows us to focus on analyzing smaller subsets of data, each containing a single sensor reading, a specific operating condition, and the health parameter (denotes as $\mathcal{D}=\{X_i,W_j,Z\}$).  While this simplification may not fully align with reality, it facilitates our goal of identifying causal relationships among $X$, $W$, and $Z$. Secondly, since the degradation effect is typically minor in the early cycles, we focus on data from later cycles (count greater than 45) where the effect of degradation is more noticeable.}

\edit{
We use DecisionTreeRegressor from sklearn with default parameters to perform regression. Since different combinations of chosen $X_i$ and $W_j$ lead to different optimal causal graphs, we report the median and mean ranking of all possible DAG structures given in Table \ref{tab:implications}.}

\begin{table}[h!]
    \centering
\begin{tabular}{c|c|c}
\hline \hline
\textbf{DAG Structure} & \textbf{Median Ranking} & \textbf{Mean Ranking} \\
\hline \hline
$Z\leftarrow[X]$ & 0.0 & 0.75 \\
$Z\leftarrow[W]$ & 1.0 & 1.14 \\
$W\leftarrow[Z]$ & 2.0 & 2.05 \\
$W\leftarrow[X]$ & 3.0 & 3.45 \\
$X\leftarrow[Z]$ & 4.0 & 3.46 \\
$X\leftarrow[W]$ & 5.0 & 5.29 \\
$Z\leftarrow[X,W]$ & 6.0 & 5.54  \\
$W\leftarrow[Z], Z\leftarrow[X]$ & 8.0 & 8.18  \\
$W\leftarrow[Z,X]$ & 8.0 & 8.86  \\
$W\leftarrow[X], Z\leftarrow[X]$ & 10.0 & 10.45  \\
$W\leftarrow[X], Z\leftarrow[W]$ & 10.0 & 11.02  \\
$X\leftarrow[Z], Z\leftarrow[W]$ & 11.0 & 10.57  \\
$X\leftarrow[Z,W]$ & 11.0 & 12.04  \\
$W\leftarrow[Z], X\leftarrow[Z]$ & 12.0 & 12.23  \\
$X\leftarrow[W], Z\leftarrow[X]$ & 13.0 & 13.70  \\
$X\leftarrow[W], Z\leftarrow[W]$ & 14.0 & 14.14  \\
$W\leftarrow[X], X\leftarrow[Z]$ & 15.0 & 15.14  \\
$W\leftarrow[Z], X\leftarrow[W]$ & 16.0 & 15.61  \\
$W\leftarrow[X], Z\leftarrow[X,W]$ & 18.0 & 18.30 \\
$W\leftarrow[Z,X], Z\leftarrow[X]$ & 19.0 & 19.70 \\
$X\leftarrow[W], Z\leftarrow[X,W]$ & 20.0 & 19.71 \\
$W\leftarrow[Z,X], X\leftarrow[Z]$ & 21.0 & 21.66 \\
\rowcolor{Gray}
$X\leftarrow[Z,W], Z\leftarrow[W]$ & \textbf{22.0} &21.11 \\
\rowcolor{Gray}
$W\leftarrow[Z], X\leftarrow[Z,W]$ &\textbf{22.0} &21.91 \\

\end{tabular}
    \caption{\edit{All possible DAG with 3 variables. Each DAG is formatted as "$\text{variable} \leftarrow [\text{cause}]$", where "variable" represents the effect variable and "cause" denotes the set of potential causal factors or parents of the variable. The median and mean ranking over all choices of $X_i$ and $W_j$ are reported.}}
    \label{tab:implications}
\end{table}

\edit{
The two most frequently observed causal structures align with our earlier reasoning, indicating that $X$ is influenced by both $W$ and $Z$, represented as $X\leftarrow[Z,W]$. It's noteworthy that the illustration provided in Figure \ref{fig:Causality}, detailed in Section \ref{sec:meth:inductive}, corresponds to $X\leftarrow[Z,W], Z\leftarrow[W]$ and is in the top two ranking. The difference between the two most frequently found causal structures is the causal directions between $W$ and $Z$. The ambiguity of the causal relationship between $W$ and $Z$ was anticipated in the C-MAPSS turbofan case study, given that the data generation process modeled degradation $Z$ as an independent process.}

\section{Reliability Type Function}
\label{Weibull}
We propose a method to determine the function $g(t)$ representing the expected Health Indicator (HI) of a system for a given cycle $t$. Inspired by reliability theory, we hypothesize that the failure time of many complex systems (i.e., $t$ for $HI(t) = 0$) follows a Weibull distribution with parameters $\beta$ and $\eta$. Additionally, we hypothesize that the time to reach any intermediate HI threshold $s$ is also Weibull distributed with parameters $\beta_s$ and $\eta_s$ \footnote{The validity of this hypothesis has been proved analytically in \cite{Pierre}.}.

Under this hypothesis, the shape parameter $\beta_s$ remains constant across different $s$ thresholds (i.e., $\beta_s = \beta \quad \forall s$.) Meanwhile, the scale parameter $\eta_s$ changes as a function of HI threshold (i.e., $\eta_s = h(s,\eta)$) \cite{Pierre}. This is because $\eta_s$ is also known as the characteristic life and corresponds to the cycles at which $63\%$ of the units have reached the threshold $s$. Since HI is decreasing as cycles increase, so does $\eta_s$. In this way, one can obtain the best-fit function $HI = h(\eta_s)$. 

 A good choice for this function (see \cite{bagdonavicius2001accelerated}) is 
\begin{equation}
\label{eta_hi}
    HI = A -(B\eta_s)^C
\end{equation}
The parameters A, B, and C are estimated from historical HI curves. Further, connecting cycle time t with HI involves the Weibull distribution's Cumulative Distribution Function, expressed as:
\begin{equation}
\label{W_cdf}
    P = 1 - exp(-(t/\eta_s)^\beta)
\end{equation}
Rewriting \ref{eta_hi} as $\eta_s = \sqrt[C]{(A - HI)}/B$ and substituting into \ref{W_cdf}, leads to

\begin{equation}
\label{HI_t}
    HI = C - ((t * (log(1 - P))^{-1/\beta})*A)^B = g(t)
\end{equation}

And thus we can obtain the best-fit HI for a fleet of units as a function of operational cycles. Note that adjusting P corresponds to adjusting the confidence of the HI. In our experiments, we have used P = 0.5. 

\section{HI Criteria}
\label{HI_criteria}

This section presents the four HI criteria used for the evaluation of the proposed method.

\begin{itemize}
    \item 
Monotonicity $M$ of health index $h_u$ of unit $u$ with $m$ observations is expressed as
{\small
\begin{align}
&M = \frac{1}{m-1} \sum_{j=1}^{m-1} |Ind(h_u^{j+1}-h_u^{j}) - Ind(h_u^{j}-h_u^{j+1}))| \\
&Ind(x) = 
\begin{cases}
  1 & x>0\\
  0 & x\leq 0
\end{cases}  \nonumber
\label{eq:monoton}
\end{align}
}

\item 
Trendability $T$ of health index $h_u$ of unit u with cycles $t_u$ is expressed as
\begin{equation}
   T = | \text{corr} (t_u,h_u)| 
\end{equation}
Where $corr(.)$ is the Spearman correlation coefficient.

\item 
Prognosability $P$ of all health indexes in a set $E^d$ is given by,
    \begin{equation}
        P = exp(-\frac{\sigma(h_u^{end})}{\mu(|h_u^{end}-h_u^{0} |)}) \quad u \in E^{\text{d}}
    \end{equation}
    Where the starting and ending HI values of unit $u$ are denoted as $h_u^{0}$ and $h_u^{end}$, respectively, while $\sigma$ and $\mu$ refer to the standard deviation and mean operators.

\item 

Mutual Information score MI between $h_u$ and $RUL_u$ for unit $u$ can be expressed as:
    {\small
    \begin{equation}
       MI = \frac{1}{m} \sum_{i=1}^{m} [1-exp(-I(h_u,RUL_u)]
    \end{equation}
}

Where $I(.)$ is the mutual information measure.

\end{itemize}




\end{document}